\documentclass[letterpaper]{article} 
\usepackage{aaai2026}  
\usepackage{times}  
\usepackage{helvet}  
\usepackage{courier}  
\usepackage[hyphens]{url}  
\usepackage{graphicx} 
\usepackage{natbib}  
\usepackage{caption} 
\usepackage{amsmath}
\usepackage{amssymb}
\usepackage{booktabs}
\usepackage{multirow}
\usepackage{tikz}
\usetikzlibrary{positioning}
\usepackage{xcolor}
\usepackage{colortbl}

\newcommand{\ours}{ICG}
\newcommand{\ourslong}{Implicit Constraint Graph}
\newcommand{\teb}{TextEditBench}

\title{Mind the Gap: Diagnosing Constraint Discovery Failures\\in Text-in-Image Editing}

\author{
    Rui Gui
}
\affiliations{
    Central South University
}

\begin{document}
\nocopyright
\maketitle

\begin{abstract}
A key challenge in multimodal reasoning is determining which visual dependencies become relevant under a specific task, rather than merely recognizing visible content.
We study this through \emph{edit-induced constraint discovery} in text-in-image editing, a controlled diagnostic setting where a local text change can activate secondary consistency constraints: given a valid editing instruction and an image, can a model identify the secondary regions that must also change?
Across 461 diagnostic cases, four MLLMs, and 19 constraint subtypes, models recover only 46\% case-level macro recall under unguided prompting versus 94\% when constraints are explicitly provided---suggesting that a substantial portion of the failure arises when models must decide which unstated dependencies to surface.
Oracle-field decomposition shows that case-specific causal explanations are the most effective partial guidance (0.782 recall), above region names (0.610) or type labels (0.646), suggesting that edit-specific causal cues account for much of the oracle gain.
A downstream experiment further shows that higher self-discovery recall does not necessarily improve task performance: unverified self-discovery introduces false positives that offset recall gains, motivating precision-aware constraint elicitation.
\end{abstract}

\section{Introduction}

Many multimodal tasks require models to surface task-relevant but unstated visual dependencies---conditions that are not mentioned in the instruction yet must be satisfied for a correct response.
Recent MLLMs excel at visual understanding~\citep{hurst2024gpt,comanici2025gemini,liu2023visual,alayrac2022flamingo}, yet they often fail to discover such dependencies spontaneously in task-driven settings.
Text-in-image editing provides a controlled diagnostic setting for this problem: a local text change can activate secondary consistency constraints elsewhere in the image (e.g., changing a discount should also update the derived price)~\citep{gui2025texteditbench}.
Our diagnostic evaluates whether MLLMs can explicitly identify edit-induced dependencies under a controlled elicitation interface, not downstream image generation.
This isolates an upstream source of failure: the same models can often identify the relevant dependency when it is made explicit, but do not surface it spontaneously.

Consider a promotional poster showing ``50\% OFF'' alongside a price tag of ``\$50.''
When instructed to change the discount to ``30\% OFF,'' an MLLM may correctly identify the need to revise the discount text but leave the price tag unchanged---despite the mathematical relationship linking them.
Similarly, changing text near a reflective surface should trigger an update to its reflection, and modifying a label on a symmetric structure should update the corresponding label on the other side.
These are \emph{implicit constraints}: conditions not stated in the editing instruction but reasonably expected to hold after the edit, arising from physical, spatial, semantic, stylistic, or cross-modal dependencies.

The relevant knowledge is nonetheless accessible: when constraints are made explicit through our oracle conditions, the same models recognize and verify 94\% of these dependencies---much of the failure appears to lie in surfacing them spontaneously when reasoning about the edit.
We call this a \emph{discovery failure}: the task of recognizing which secondary elements are affected by a given edit (\emph{edit-induced constraint discovery}) is not triggered, and models default to specifying only the local text replacement without considering broader consequences.

Work on latent knowledge~\citep{burns2022discovering,yan2025hidden} suggests models can hold information they do not spontaneously express. This prior work motivates our diagnostic framing; the empirical claims below rest on the measured discovery--verification gap.
The measured gap is also heterogeneous: unguided recall varies $2.7\times$ across constraint categories (and $6.3\times$ across the finer subtypes).

\begin{figure}[t]
\centering
\includegraphics[width=0.95\columnwidth]{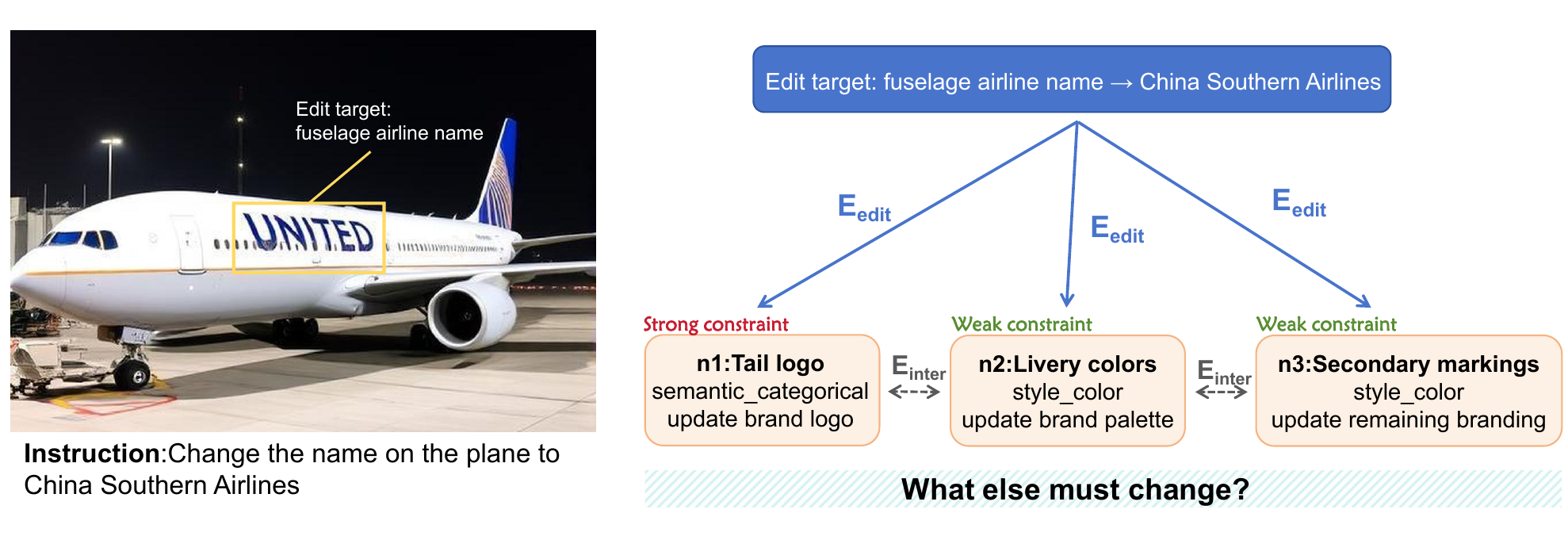}
\caption{Example \ours{} annotation for an aircraft rebranding edit. Changing the fuselage airline name triggers three constraint nodes spanning two categories (semantic and style). The tail logo ($n_1$) must match the new brand identity, the livery colors ($n_2$) must adopt the target airline's palette, and secondary markings ($n_3$) must maintain visual coherence. Edges indicate edit-target-to-region dependencies and optional consistency relations among affected regions.}
\label{fig:icg_example}
\end{figure}

To study this, we introduce the \emph{\ourslong{}} (\ours{}), a structured annotation schema that makes explicit the edit-induced constraints triggered by an editing instruction (Figure~\ref{fig:icg_example}).
Unlike scene graphs~\citep{johnson2015image,krishna2017visual,zellers2018neural,chang2021comprehensive} that capture pre-existing objects, attributes, and relationships, or visual grounding approaches~\citep{yang2023set,peng2024grounding} that localize existing image regions, \ours{} represents \emph{editing-induced} constraints: how changing element $A$ should cascade to elements $B$ and $C$.
Each constraint node specifies an affected region, its dependency reason, constraint type, and expected update; optional inter-node edges capture consistency requirements among affected regions.

We design a controlled diagnostic framework with seven prompting conditions (Figure~\ref{fig:pipeline}) spanning no guidance (Direct) to full oracle provision. Adversarial, $E_\text{inter}$ ablation, and oracle-field decomposition controls then test whether the oracle gain reflects case-specific causal constraint information, inter-node consistency edges, or blind compliance.

We evaluate 461 cases across four strong contemporary MLLMs, covering 19 constraint subtypes in five categories: Physical, Spatial, Semantic, Style, and Cross-modal.
Our experiments yield three main findings:
\begin{enumerate}
    \item \textbf{Discovery failure is structured.} Unguided recall varies $2.7\times$ across constraint categories, from cross-modal to semantic constraints; finer subtype-level variation ($6.3\times$) is larger but exploratory.
    \item \textbf{Case-specific causal information is the strongest partial oracle cue.} The $E_\text{inter}$ ablation ($\Delta = +0.007$) and adversarial control (90.5\% rejection) argue against inter-node edges or blind compliance as primary explanations. By contrast, oracle-field decomposition shows that de-identified causal explanations (0.782 recall) outperform region names (0.610) and type labels (0.646), indicating that edit-specific dependency reasons account for a large part of the oracle guidance.
    \item \textbf{Higher discovery recall does not automatically improve downstream use.} A downstream experiment shows that NL-Self's +0.155 recall gain over Direct yields no plan-satisfaction improvement ($d = -0.05$, $p = 0.62$), because unverified self-discovery introduces false positives that offset recall gains.

\end{enumerate}

\noindent In summary, we contribute:
(1) an edit-induced constraint discovery formulation and structured annotation schema (19 subtypes, 5 categories);
(2) a controlled diagnostic framework with seven prompting conditions plus oracle-field decomposition (Region/Reason/Type-Only), separating discovery from verification;
(3) an empirical characterization of where discovery fails and which oracle information helps, with adversarial, format, and decomposition controls;
and (4) evidence that higher discovery recall alone does not improve downstream use, motivating precision-aware constraint elicitation.

\section{Related Work}

\paragraph{Text-in-Image Editing and Evaluation.}
Diffusion-based methods improve text rendering and local text editing~\citep{tuo2024anytext,chen2023textdiffuser,yang2023glyphcontrol}, while instruction-based and MLLM-guided editors use natural-language or multimodal guidance to produce better edited images~\citep{brooks2023instructpix2pix,zhang2023magicbrush,basu2023editval,fu2024guiding,huang2024smartedit,yin2025reasonedit}. 
Editing benchmarks evaluate final outputs under fidelity, instruction following, preservation, or reasoning-aware criteria~\citep{zhao2026envisioning,liu2026grade,yosef2025editinspector,ma2024i2ebench,yi2026vde,ding2026edit,lin2026worldedit,wang2025complexbench}.
In contrast, we use editing instructions as controlled probes: our object of evaluation is not the edited image, but whether an MLLM can explicitly identify unmentioned visual dependencies activated by an instruction-image pair.

\paragraph{Structured Prompting and Visual Dependency Reasoning.}
CoT and decomposed or graph-structured prompting can improve reasoning but have mixed effects in visual tasks~\citep{wei2022chain,zhang2023multimodal,shao2024visual,izadi2026visual,khot2022decomposed,besta2024graph}; visual grounding and scene representations localize or encode existing objects and relations~\citep{yang2023set,peng2024grounding,johnson2015image,krishna2017visual,zellers2018neural,chang2021comprehensive,shen2025vcd}.
Our setting differs because the relevant dependencies are instruction-conditioned: a secondary region matters only because a particular edit would make it inconsistent with the target.

\paragraph{Implicit Reasoning and Problem Detection.}
Work on latent knowledge and elicitation shows that model behavior can change with how information is queried~\citep{burns2022discovering}. Most closely related, HiPS~\citep{yan2025hidden} shows that MLLMs often comply with flawed instructions rather than surface hidden problems. Our instructions are valid; the challenge is instead to identify secondary dependencies that the request implicitly activates. Related failures in semantic fixation and constraint satisfaction further motivate studying when models fail to surface task-relevant information~\citep{alam2026beyond,kovalev2025craft,huang2023t2i}.

\section{Method}

\subsection{Problem Formulation}

Given an image $I$ and an editing instruction $T$, an implicit constraint is a condition $c$ that is (1) not explicitly stated in $T$, (2) arises from visual, semantic, physical, or stylistic dependencies between the edited element and other elements in $I$, and (3) would reasonably be expected to hold after the edit.
Our goal is to \emph{diagnose} whether an MLLM can identify all implicit constraints triggered by $(I, T)$.

\subsection{Edit-Induced Constraint Annotation Schema}

We define a structured annotation schema for edit-induced constraints, represented as $G(V, E_\text{edit}, E_\text{inter})$:
\begin{itemize}
    \item $V = \{n_1, \ldots, n_k\}$: Constraint nodes, each representing a region implicitly affected by the edit. Each $n_i$ carries attributes: spatial region $r_i$, constraint type $\tau_i \in \mathcal{T}$, expected update $u_i$, and dependency reason $d_i$ (why this region is affected).
    \item $E_\text{edit}$: Edges from the edit target to each affected node, encoding the edit-induced dependency.
    \item $E_\text{inter}$: Optional edges capturing consistency requirements \emph{between} affected nodes.
\end{itemize}
This representation separates four pieces of diagnostic information: the affected region, the constraint type, the expected update, and the reason the edit propagates there. Optional inter-node edges record consistency relations among affected regions.

\subsection{Constraint Taxonomy}

We organize implicit constraints into 5 categories with 19 subtypes (Table~\ref{tab:taxonomy}).
Categories are informed by scene graph modeling~\citep{krishna2017visual}, controllable generation~\citep{zhang2023adding}, and perceptual organization theory~\citep{treisman1980feature,greff2020binding}; subtypes were stabilized through iterative pilot annotation until no new type appeared in five consecutive cases (details in supplementary).
The taxonomy was finalized \emph{before} case selection from non-TextEditBench sources (Pexels, Unsplash, Canva), reducing circularity for the newly collected portion; TextEditBench overlap is addressed by source-stratified analyses in the supplementary material.
The full taxonomy with descriptions and examples appears in the supplementary material.
\begin{table}[t]
\centering
\small
\caption{Constraint taxonomy: 5 categories, 19 subtypes (condensed for space; see the supplementary material for full table). $N$ = constraint nodes across 461 cases.}
\label{tab:taxonomy}
\begin{tabular}{@{}llr@{}}
\toprule
\textbf{Category} & \textbf{Subtype} & \textbf{$N$} \\
\midrule
\multirow{2}{*}{Physical} & Reflection / Shadow & 79 \\
 & Perspective / Occlusion & 89 \\
\midrule
\multirow{2}{*}{Spatial} & Mirror / Symmetry & 32 \\
 & Repetition & 24 \\
\midrule
\multirow{3}{*}{Semantic} & Numerical & 19 \\
 & Categorical / Functional & 75 \\
 & Promotional / Visual-Data & 72 \\
\midrule
Style & Font / Color / Layout / Material & 173 \\
\midrule
Cross-modal & Scene / Temporal / Cultural & 213 \\
\midrule
\multicolumn{2}{l}{\textbf{Total}} & \textbf{776} \\
\bottomrule
\end{tabular}
\end{table}

\begin{figure*}[t]
\centering
\includegraphics[width=0.95\textwidth]{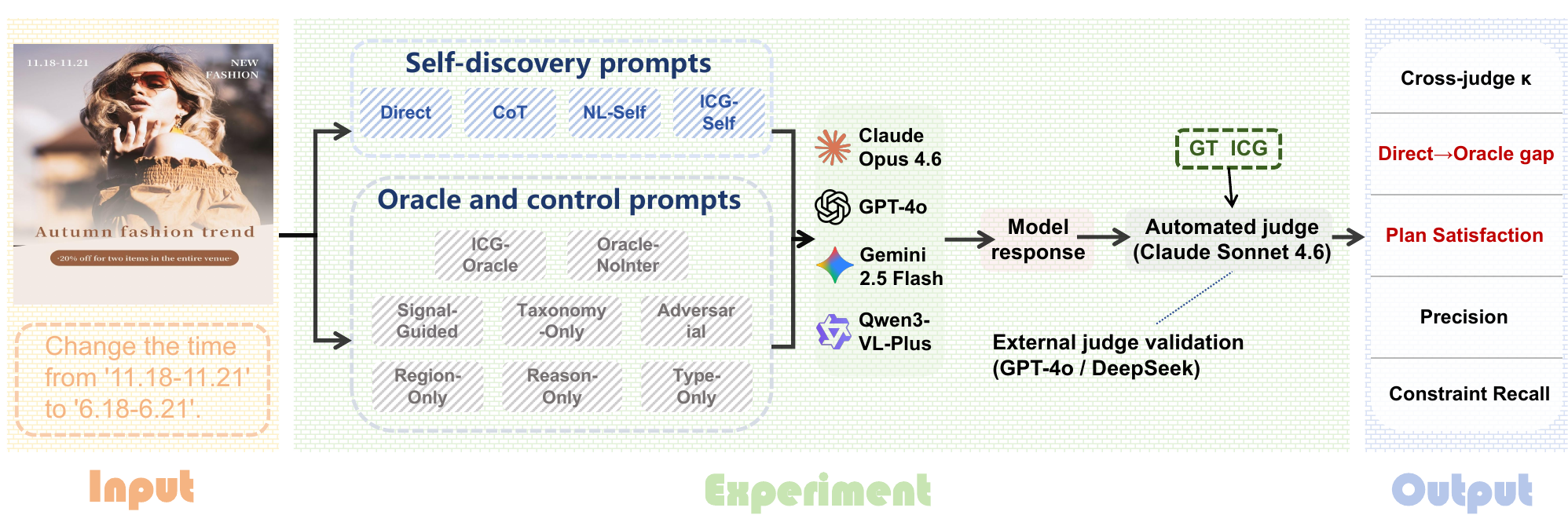}
\caption{Diagnostic framework overview. Each case (image + editing instruction) is evaluated under seven prompting conditions spanning no guidance to full oracle. Model responses are scored against the ground-truth \ours{} by an automated judge, yielding per-constraint recall and precision. Two ablation conditions (Adversarial, Oracle-NoInter) and three oracle-field decomposition conditions (Region-Only, Reason-Only, Type-Only) provide additional control evidence. A downstream planning evaluation further tests whether discovery improvements translate into better edit plans (plan satisfaction).}
\label{fig:pipeline}
\end{figure*}

\subsection{Diagnostic Pipeline}

\textbf{Stage 1: Constraint Elicitation.}
Given $(I, T)$, the model identifies implicit constraints. Under \ours{}-Self, the model outputs a structured JSON graph; under \ours{}-Oracle, the human-annotated graph is provided.
\textbf{Stage 2: Response Formation.}
The model reports discovered constraints or, under oracle conditions, uses provided constraints to formulate the required updates.
\textbf{Stage 3: Evaluation.}
An automated judge compares the model's output against the ground-truth \ours{} to measure constraint recall.
We focus on Stage 1 (constraint discovery) as the primary diagnostic target.
This pipeline is an evaluation decomposition for diagnosing constraint discovery; it does not assume that real editing systems have an explicit planning stage with this architecture.

\textbf{Adversarial Oracle Control.}
To measure acceptance bias under oracle prompting, we inject one plausible but factually incorrect constraint per case alongside the ground-truth constraints.
Models must label each constraint as VALID or INVALID.

\textbf{$E_\text{inter}$ Ablation.}
We compare \ours{}-Oracle (full graph) against Oracle-NoInter (same nodes and $E_\text{edit}$, with only $E_\text{inter}$ removed) on the 228 cases containing inter-constraint dependencies.

\section{Experimental Setup}

\subsection{Data Construction}

We curate 461 diagnostic cases, each consisting of an image $I$, an editing instruction $T$, and a human-annotated ground-truth \ours{} with 776 total constraint nodes (1.68 per case on average).
Images are sourced from \teb{}~\citep{gui2025texteditbench}, Pexels, Unsplash, and Canva (filtered by automated pre-screening). Underrepresented subtypes were supplemented with additional cases collected from other sources to ensure coverage.
Candidate selection uses a three-stage protocol: (1) automated pre-screening by Claude Opus 4.6; (2) human inspection of visual quality; (3) final review of complete annotations.
Each constraint node must have its affected region visibly present in the image and be based on direct observation rather than assumption.

\textbf{Annotation quality.}
Three annotators with computer-science backgrounds independently classified a stratified sample of 60 nodes: Fleiss' $\kappa = 0.77$ (substantial; details in the supplementary material).
An independent verification audit on a $20\%$ stratified sample of all 776 nodes (156 nodes) checks existence, region correctness, update correctness, and reason validity; per-dimension agreement ranges from 84.6\% to 90.4\% across four dimensions (details in the supplementary material). This audit complements the subtype-classification agreement above by showing that the affected region, expected update, and dependency reason are also largely supported by an independent verifier.

\subsection{Models}

We evaluate Claude Opus 4.6, GPT-4o~\citep{hurst2024gpt}, Gemini 2.5 Flash~\citep{comanici2025gemini}, and Qwen3-VL-Plus~\citep{bai2025qwen3}---spanning four major commercial model families (Anthropic, OpenAI, Google, Alibaba).
All use temperature $= 0$ and \texttt{max\_tokens} $= 4096$ (Gemini: 16384 under CoT). Images standardized to 1024px max long edge. The automated judge is Claude Sonnet 4.6; adversarial constraints are generated by GPT-4o for independence from the Claude judge family.

\subsection{Experimental Conditions}

We test each case under seven prompting conditions (Figure~\ref{fig:pipeline}):
\textbf{Direct} (no guidance),
\textbf{CoT} (step-by-step),
\textbf{\ours{}-Self} (JSON graph output),
\textbf{NL-Self} (free-form, same information dimensions),
\textbf{Signal-Guided} (a free-form prompt that directs models to inspect common high-risk cues such as repeated text, reflections, shadows, prices, charts, and brand/style consistency, without providing case-specific affected regions),
\textbf{Taxonomy-Only} (19-subtype checklist, no case-specific nodes), and
\textbf{\ours{}-Oracle} (ground-truth constraints provided);
plus \textbf{\ours{}-Adversarial} (oracle + one injected invalid constraint), \textbf{Oracle-NoInter} (oracle without $E_\text{inter}$ edges), and three oracle-field decomposition conditions: \textbf{Region-Only} (affected regions without type or reason), \textbf{Reason-Only} (de-identified causal explanations without region names), and \textbf{Type-Only} (constraint type labels only).
Full prompts appear in the supplementary material.

\subsection{Evaluation}

An automated judge (Claude Sonnet 4.6) evaluates whether each ground-truth constraint was identified, requiring the response to identify the affected region, state that it needs to be updated, and give a dependency reason matching the ground-truth constraint.
In this work, ``discovery'' refers not merely to naming an affected region, but to producing a minimally actionable constraint: the region, the need for an update, and the causal link to the edit.
We report \textbf{Constraint Recall} = $|\text{found}| / |\text{GT}|$ as the primary metric, since our diagnostic question is whether models \emph{discover} ground-truth constraints---a recall-oriented task by design.
Precision analysis on self-generated constraints (87.3\% ICG-Self, 83.1\% NL-Self) is reported in the Results section alongside recall; details in the supplementary.

\textbf{Cross-judge validation.}
Because LLM-as-judge evaluations carry known biases including self-enhancement and verbosity preferences~\citep{zheng2023judging}, we validate the headline pattern with two additional judges, including one fully out-of-pool judge.
GPT-4o judging preserves the Direct$\rightarrow$Oracle gap under macro-averaged scoring ($+0.578$). DeepSeek-V4-flash, a fully out-of-pool judge, also preserves the gap under node-level scoring ($+0.472$; primary Claude on the same table: $+0.475$), with substantial agreement on the core conditions ($\kappa = 0.675$ and approximately $0.60$--$0.70$).
Human validation and guided-condition checks in the supplementary show the same qualitative pattern, although Taxonomy-Only is stricter under DeepSeek.

\section{Results and Analysis}

\subsection{Main Results}

Table~\ref{tab:main_results} presents constraint recall across models and conditions (see Figure~\ref{fig:pipeline} for the overall evaluation pipeline).
Unless otherwise noted, main results use macro-averaged case-level recall; per-category analyses (Table~\ref{tab:per_type}) use micro-averaged node-level recall, which explains the slight difference between overall Oracle values (0.938 vs.\ 0.921).

\begin{table}[t]
\centering
\small
\caption{Constraint Recall. ICG-S = ICG-Self (JSON); NL-S = NL-Self (free-form); SG = Signal-Guided; Tax. = Taxonomy-Only; ICG-O = ICG-Oracle. Bold = best among Direct, CoT, ICG-S, NL-S, and SG. Taxonomy-Only (gray) provides an exhaustive generic checklist and is analyzed separately.}
\label{tab:main_results}
\setlength{\tabcolsep}{3pt}
\begin{tabular}{@{}lcccccc c@{}}
\toprule
\textbf{Model} & \textbf{Direct} & \textbf{CoT} & \textbf{ICG-S} & \textbf{NL-S} & \textbf{SG} & \textbf{Tax.} & \textbf{ICG-O} \\
\midrule
Claude & 0.659 & 0.615 & 0.648 & \textbf{0.728} & 0.616 & \textcolor{gray}{0.889} & 0.916 \\
GPT-4o & 0.320 & 0.486 & 0.406 & 0.587 & \textbf{0.589} & \textcolor{gray}{0.736} & 0.976 \\
Gemini & 0.455 & 0.523 & 0.558 & 0.599 & \textbf{0.710} & \textcolor{gray}{0.662} & 0.931 \\
Qwen3 & 0.412 & 0.464 & 0.521 & 0.549 & \textbf{0.626} & \textcolor{gray}{0.744} & 0.927 \\
\midrule
\textbf{All} & 0.461 & 0.522 & 0.534 & 0.616 & \textbf{0.635} & \textcolor{gray}{0.758} & 0.938 \\
\bottomrule
\end{tabular}
\end{table}

\textbf{The discovery gap.}
Direct, CoT, and \ours{}-Self remain substantially lower than \ours{}-Oracle: 0.461, 0.522, and 0.534 versus 0.938, an approximately +0.476 Direct$\rightarrow$Oracle gap.\footnote{McNemar tests: Direct$\rightarrow$Oracle $\chi^2 = 1344$, $p < 10^{-15}$; Direct$\rightarrow$CoT $\chi^2 = 67.0$, $p < 10^{-15}$; CoT$\rightarrow$ICG-Self $p = 1.0$ (symmetric discordant pairs). GEE logistic regression shows that CoT, ICG-Self, and Oracle each improve over Direct after accounting for within-case clustering (details in the supplementary material).}
Models recover 93.8\% case-level macro recall (92.1\% node-level micro recall) when explicitly provided but achieve only 46.1\% macro recall (44.7\% micro recall) under unguided prompting---models often focus on the requested local text replacement while omitting secondary consistency constraints.
Three human annotators on a 100-case stratified subset reach mean micro-averaged Direct recall 0.67 (individual: 0.64, 0.68, 0.70)---above unguided models (0.447 micro) but below Oracle (0.921 micro). The constraints are therefore identifiable but nontrivial.

\textbf{Format and task framing both matter.}
NL-Self differs from \ours{}-Self in two ways: it allows free-form output and more directly asks the model to search for edit-induced dependencies.
To isolate the first factor, we run a strict format ablation that keeps the task fixed and varies only the required output format; free-form output exceeds JSON by +0.042 macro recall (+0.038 paired node-level recall, $p < 0.001$; details in the supplementary material).
The format penalty is model-dependent: GPT-4o shows the largest effect ($+0.087$ macro) while Claude shows almost none ($+0.010$), suggesting that sensitivity to output structure varies across model families.
This explains part, but not all, of the NL-Self vs.\ \ours{}-Self gap (+0.082), consistent with an additional effect of explicit dependency-search framing.
NL-Self outperforms Direct by +0.155, showing that explicit dependency-search prompts recover constraints missed by unguided prompting.

\textbf{Generic taxonomies are helpful but insufficient.}
Taxonomy-Only provides a generic 19-subtype checklist without case-specific affected regions or dependency reasons; we analyze it separately from the best non-oracle marking in Table~\ref{tab:main_results}.
It improves over unguided prompting under the primary judge, but under the stricter out-of-pool DeepSeek judge it remains far below Oracle (details in the supplementary material).
Generic reminders help models search more broadly but do not solve constraint discovery: the model must still determine which constraints a particular edit causally triggers in a particular image.

\textbf{Signal-Guided improves over open-ended self-discovery for most models.}
Signal-Guided (SG) provides compact visual search cues---reflections, repeated text, numerical relationships, style coherence---without case-specific affected regions or a full 19-subtype checklist.
For GPT-4o, Gemini, and Qwen3, SG matches or exceeds NL-Self among conditions that do not provide an exhaustive generic checklist (0.589, 0.710, 0.626).
Claude is the exception: its NL-Self (0.728) exceeds SG (0.616), consistent with Claude's stronger baseline ability to independently search for dependencies without external cues.
The pattern suggests that prioritized search cues can be more effective than open-ended self-discovery when spontaneous discovery is weaker.

\textbf{OCR-assisted prompting helps but does not close the gap.}
On a stratified 50-case subset, explicit text localization improves recall over Direct ($\Delta = +0.274$) but still leaves a substantial gap to Oracle ($\Delta = -0.235$), showing that visual search contributes to the gap but does not by itself solve edit-induced constraint discovery.
The benefit is also model-dependent: GPT-4o gains the most ($+0.393$), consistent with its low Direct recall on visually complex text, while Claude gains the least ($+0.113$), consistent with its already-stronger baseline visual text reading (details in the supplementary material).

\textbf{Oracle compresses inter-model differences.}
Under Direct, recall spans 0.320--0.659 (0.339 spread); under Oracle, 0.916--0.976 (0.060 spread)---much of the observed variation across models appears in the unguided discovery setting and shrinks once constraints are made explicit.
Oracle recall is best viewed as a high-opportunity guided condition: the prompt simultaneously reduces discovery burden, visual search effort, and grounding ambiguity.
Nonetheless, full Oracle does not eliminate all errors: residual failure remains at 6.2\% under case-level macro averaging and 7.9\% under node-level micro averaging. These errors are concentrated in style constraints (Oracle 0.883), suggesting that some failures reflect reasoning or update-specification limits even when constraints are provided; see the supplementary material for failure source decomposition.

\textbf{Self-generated constraints are valid but incomplete.}
ICG-Self and NL-Self have high prediction precision (87.3\% and 83.1\%; supplementary), so the discovery gap is driven mainly by \emph{omission} rather than hallucination.
However, the remaining false positives---though a minority---still matter for downstream use: as the downstream experiment below shows, even a modest false-positive rate can introduce unnecessary edit steps that offset recall gains.

\subsection{Per-Constraint-Type Analysis}

\begin{figure}[t]
\centering
\includegraphics[width=0.95\columnwidth]{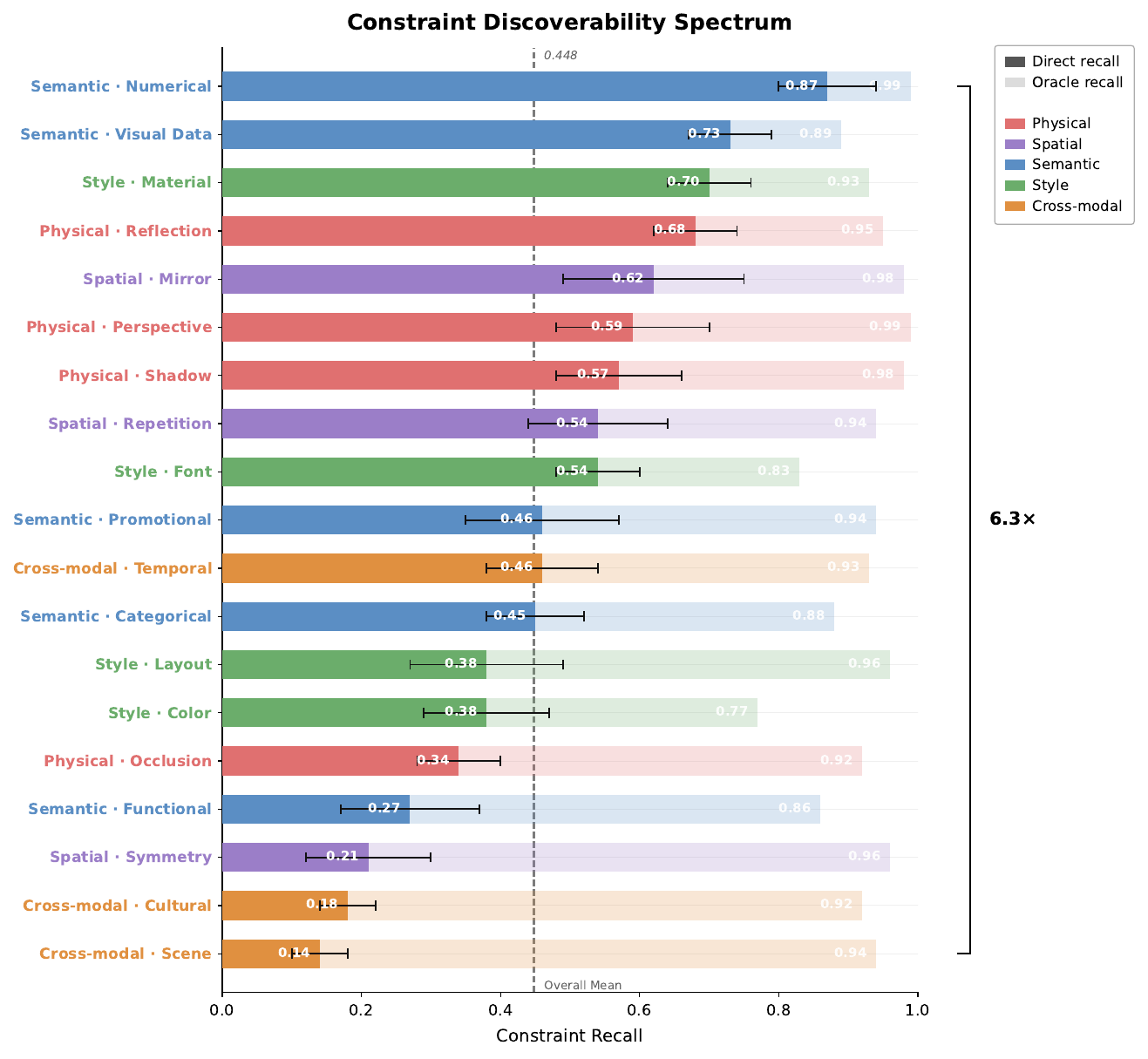}
\caption{Constraint discoverability spectrum: per-subtype Direct recall (sorted) with 95\% bootstrap CIs ($B = 2{,}000$) and Oracle recall. All 19 subtypes have $N \geq 14$; unguided discovery rates span 0.14--0.87 (6.3$\times$).}
\label{fig:subtype_recall}
\end{figure}

Table~\ref{tab:per_type} presents micro-averaged recall by category.
Cross-modal constraints show the largest Oracle delta (+0.716); semantic constraints show the smallest (+0.334).
Notably, CoT \emph{reduces recall} for Physical and Style categories in our setting, while substantially helping Spatial, Semantic, and Cross-modal constraints. One possible explanation is that Physical and Style constraints rely on fine-grained visual feature binding (shadow shapes, color palettes) where step-by-step verbal reasoning adds noise~\citep{izadi2026visual}, whereas Semantic and Cross-modal constraints benefit from the logical decomposition that CoT encourages.

\begin{table}[t]
\centering
\small
\caption{Micro-averaged Constraint Recall by category (all models, 461 cases). Node-level micro-averaged recall; differs from Table~\ref{tab:main_results}'s case-level macro-averaged recall of 0.938 due to aggregation method. $\Delta$ = Oracle minus Direct. Parentheses = node count.}
\label{tab:per_type}
\begin{tabular}{@{}lccccc@{}}
\toprule
\textbf{Category} & \textbf{Dir.} & \textbf{CoT} & \textbf{ICG-S} & \textbf{ICG-O} & \textbf{$\Delta$} \\
\midrule
Physical (168) & 0.519 & 0.440 & 0.476 & 0.963 & +0.443 \\
Spatial (56) & 0.455 & 0.705 & 0.701 & 0.955 & +0.500 \\
Semantic (166) & 0.566 & 0.741 & 0.705 & 0.901 & +0.334 \\
Style (173) & 0.548 & 0.439 & 0.540 & 0.883 & +0.335 \\
Cross-modal (213) & 0.211 & 0.454 & 0.372 & 0.927 & +0.716 \\
\midrule
\textbf{Overall (776)} & 0.447 & 0.527 & 0.527 & 0.921 & +0.475 \\
\bottomrule
\end{tabular}
\end{table}

At the category level (Table~\ref{tab:per_type}, all cells $N \geq 56$), unguided discovery varies $2.7\times$, from cross-modal (0.21) to semantic (0.57); this coarse pattern is statistically robust.
At the finer subtype level (Figure~\ref{fig:subtype_recall}), recall spans 0.14 (cross-modal scene) to 0.87 (numerical)---a $6.3\times$ range.
The spectrum endpoints are robust. Adjacent mid-spectrum subtype rankings remain exploratory because several subtypes have $N \leq 21$ and overlapping confidence intervals.
A plausible explanation is that discoverability depends on two interacting factors: \emph{edit signal salience} (whether the edit itself cues a dependency) and \emph{relation inferability} (whether the dependency can be derived without external knowledge or cross-region search).
For example, numerical constraints (Direct recall 0.87) are highly discoverable because changing a number immediately signals that derived quantities may be affected, and the mathematical relation is directly inferable from the edit.
At the other extreme, cross-modal scene constraints (Direct recall 0.14) are rarely found because decorative text changes carry no obvious cue that the visual scene context may be violated, and detecting the mismatch requires contextual reasoning about what the scene depicts.
A structural complexity coding that ranks subtypes by the indirectness of the edit-to-constraint relation correlates with Direct recall ($\rho = -0.53$, $p = 0.020$; details in the supplementary material), though this is exploratory.

\paragraph{Model-Specific Patterns.}
Claude shows the highest Direct recall (0.659) but CoT slightly \emph{decreases} its performance ($-0.044$), consistent with the category-level pattern above; GPT-4o benefits most from CoT (+0.166).
Across models, NL-Self is the most consistent improvement over Direct.
Qualitative case studies spanning these discovery regimes appear in the supplementary material.

\subsection{Adversarial Oracle Validation}
\label{sec:adversarial}

Across 1,844 evaluations (461 cases $\times$ 4 models), models correctly reject 90.5\% of injected invalid constraints (per-model range 84.2--95.0\%; breakdown in the supplementary material).
The per-model variation is informative: Claude rejects 95.0\% while GPT-4o rejects only 84.2\%, suggesting that acceptance bias varies across model families and is not a uniform property of oracle prompting.
A human annotator rated 87\% of a 54-item sample as well-designed adversarials; on these, rejection is 91.5\%. This supports the narrower conclusion that the high Oracle score is not driven mainly by uncritical acceptance of provided constraints.

\subsection{$E_\text{inter}$ Ablation}
\label{sec:einter}

Comparing Full Oracle with Oracle-NoInter on the 228 cases containing inter-constraint edges yields an overall difference of only $\Delta = +0.007$ (macro recall), with no consistent direction across models---Claude and GPT-4o do marginally better with $E_\text{inter}$ edges while Gemini and Qwen3 do marginally better without them (full per-model results in the supplementary material).
Under our annotation scheme, the oracle gain is better explained by \emph{node-level content}---particularly dependency reasons---than by the optional consistency links among affected nodes.
These links may still matter in more complex editing scenarios (e.g., multi-hop constraint propagation), but for our cases the strongest diagnostic signal resides in individual node attributes.

\subsection{Oracle Decomposition: Which Oracle Fields Help Most?}
\label{sec:oracle_decomp}
To further isolate which oracle components drive the performance boost, we run three ablation conditions on all 461 cases:
\textbf{Region-Only} (which regions are affected, but not why or what type),
\textbf{Reason-Only} (abstract causal explanations of why each region is affected, with all region names, object references, and specific text content removed to reduce information leakage---models must locate affected regions themselves), and
\textbf{Type-Only} (constraint type labels like ``Semantic\_Numerical,'' but not which region or why).
Table~\ref{tab:oracle_decomp} shows the results.

\begin{table}[t]
\centering
\small
\caption{Oracle decomposition (461 cases $\times$ 4 models). Each condition provides one oracle component.}
\label{tab:oracle_decomp}
\begin{tabular}{@{}lcccc@{}}
\toprule
\textbf{Model} & \textbf{Region} & \textbf{Type} & \textbf{Reason} & \textbf{Full} \\
\midrule
Claude Opus 4.6 & 0.739 & 0.848 & \textbf{0.923} & 0.916 \\
GPT-4o & 0.511 & 0.547 & 0.775 & 0.976 \\
Gemini 2.5 Flash & 0.645 & 0.639 & 0.723 & 0.931 \\
Qwen3-VL-Plus & 0.544 & 0.550 & 0.705 & 0.927 \\
\midrule
\textbf{All} & 0.610 & 0.646 & \textbf{0.782} & 0.938 \\
\bottomrule
\end{tabular}
\end{table}

\textbf{Reason-Only substantially outperforms Region-Only and Type-Only.}
Reason-Only achieves 0.782 recall, above Region-Only (0.610, $\Delta = +0.172$) and Type-Only (0.646, $\Delta = +0.136$).
This ranking addresses what kind of oracle information helps most.
Region-Only tells the model \emph{where} to look; Reason-Only tells it \emph{why} a dependency exists. Knowing why outperforms knowing where, suggesting that recognizing the causal link is a major bottleneck---though Reason-Only may also implicitly convey type information and visual search cues.
The Reason-vs-Type comparison is less clean: a causal reason can implicitly signal the constraint type, so part of the Reason-Only advantage may overlap with what Type-Only already provides.
However, Reason-Only (0.782) still falls short of Full Oracle (0.938). This remaining gap is consistent with cases where causal descriptions alone are insufficient---models may additionally need explicit region localization or update-specification information, particularly for style and spatial constraints where the causal description does not uniquely determine the affected visual element.
One anomaly: Claude achieves higher recall under Reason-Only (0.923) than under Full Oracle (0.916); we treat this as model-specific prompt variability near ceiling, since all other models and the pooled average still favor Full Oracle.
Residual leakage from de-identification is quantified in the supplementary material (19\% exact region recovery from de-identified reasons vs.\ 52\% from raw reasons).

\paragraph{Synthesis.}
These controls narrow the source of the Oracle gain: inter-node consistency edges and uncritical compliance explain little of the gap, while generic type awareness helps but is less effective than case-specific causal information. The largest partial gain comes from supplying the causal relation linking the edit to affected regions.

\section{Discussion}

\paragraph{From Discovery to Downstream Use: Recall Gains Require Precision.}
A natural assumption is that helping models discover more constraints will directly improve edit plans. We test this with a planning experiment (60 stratified cases, 100 constraints, 3 conditions; protocol in the supplementary material).
Providing ground-truth constraints yields near-perfect plan satisfaction (0.990); unguided Direct prompting drops to 0.700 ($\Delta = 0.29$, $p < 10^{-8}$), confirming that missed constraints degrade planning.
Crucially, however, NL-Self---which achieves +0.155 higher discovery recall than Direct---shows \emph{no} improvement in plan satisfaction (0.680 vs.\ 0.700; $\Delta = -0.02$, Cohen's $d = -0.05$, $p = 0.62$). The effect, if anything, is slightly negative.
This dissociation arises because planning quality depends on \emph{precision} as well as recall: NL-Self's broader search surfaces additional true constraints but also introduces false positives that generate unnecessary edit steps, offsetting the recall benefit.
A two-stage discover-then-verify variant also remains well short of Oracle (0.712), suggesting that post-hoc filtering alone may not compensate for unreliable initial discovery.\footnote{The two-stage pipeline reaches 0.712 plan satisfaction on a related 58-case run after filtering 31.2\% of candidate constraints; full setup and comparability caveat are in the supplementary material.}
This motivates precision-aware constraint elicitation: closing the discovery gap requires not just broader search but surfacing the right constraints without flooding the plan with spurious ones.

\paragraph{Implications for Multimodal Reasoning and Editing Systems.}
We interpret this pattern as a knowledge-deployment gap in this setting: models can often apply or verify the relevant relational knowledge once it is explicitly cued (evidenced by high Oracle recall) but do not reliably activate it from the editing instruction alone. This gap may be relevant to other multimodal tasks requiring task-activated dependency reasoning, although our evidence is specific to text-in-image editing.
For editing-oriented applications, these results motivate explicit dependency elicitation as a useful diagnostic or auxiliary step: a model first proposes candidate secondary changes grounded in visible regions and dependency reasons, which can then be verified before use.

\paragraph{Cross-Family Generalization.}
A preliminary open-weight check (GLM-4.6V on a 50-case stratified subset) shows the same qualitative pattern: Direct recall 0.388 rises to Oracle recall 0.970 (gap $+0.582$), within the proprietary range. A single model on 50 cases is suggestive rather than conclusive.

\paragraph{Limitations.}
\label{sec:limitations}
Our 461 cases cover all 19 subtypes with $N \geq 14$, but several remain below 25, limiting per-subtype ranking stability.
The primary judge (Claude Sonnet 4.6) shares lineage with one evaluated model, but two additional judges---GPT-4o ($\kappa = 0.617$--$0.646$) and out-of-pool DeepSeek ($\kappa = 0.598$--$0.698$)---both reproduce the Direct$\rightarrow$Oracle gap. Same-family judging is therefore a mitigated source of uncertainty rather than a driver of the headline result.
Candidate pre-screening used Claude Opus 4.6; a source-stratified analysis shows Claude's Direct recall is actually \emph{lower} on pre-screened cases (0.590) than on other sources (0.679--0.681), providing no support for a pre-screening bias.
The downstream experiment evaluates text-based edit plans rather than final generated images; end-to-end generation evaluation remains future work.
Our study is a behavioral diagnostic of language-mediated constraint discovery; it does not claim to reveal the internal latent representations used by end-to-end generation models.

\section{Conclusion}

We introduced edit-induced constraint discovery as a diagnostic task for task-activated visual dependency reasoning, instantiated in text-in-image editing. Across 461 cases and four MLLMs, models achieve less than half of secondary consistency recall under unguided prompting but recover most constraints when those dependencies are made explicit. The gap is not uniform: it varies by constraint type and is largest for dependencies requiring weak edit cues, cross-region search, or external knowledge.

Oracle-field decomposition shows that case-specific causal descriptions provide the strongest partial guidance, while a downstream experiment shows that recall gains alone are insufficient when self-discovery also introduces false positives. These results suggest that task-activated dependency discovery is a distinct bottleneck in multimodal reasoning, and that effective constraint elicitation requires precision as well as recall.

\clearpage
\bibliography{aaai2026}

\end{document}


\nocopyright
\maketitle

\appendix

\section*{Overview}

This supplementary provides:
\begin{itemize}
\item Appendix~\ref{app:prompts}: Full prompts for all experimental conditions (including NL-Self, Signal-Guided, and oracle decomposition conditions Region-Only, Reason-Only, Type-Only).
    \item Appendix~\ref{app:judge}: Automated judge prompt and evaluation criteria.
    \item Appendix~\ref{app:taxonomy_full}: Complete constraint taxonomy with descriptions.
    \item Appendix~\ref{app:subtype_full}: Full per-subtype recall results.
    \item Appendix~\ref{app:einter}: $E_\text{inter}$ ablation details.
    \item Appendix~\ref{app:oracle_model}: Oracle recall by category and model.
    \item Appendix~\ref{app:oracle_decomp}: Oracle decomposition experiment details.
    \item Appendix~\ref{app:failure}: Failure source decomposition.
    \item Appendix~\ref{app:adversarial}: Adversarial constraint generation.
    \item Appendix~\ref{app:format_ablation}: Strict format ablation (minimal pair).
    \item Appendix~\ref{app:case_examples}: Qualitative case studies.
    \item Appendix~\ref{app:statistical_robustness}: Statistical robustness analysis (GEE, cluster bootstrap).
    \item Appendix~\ref{app:judge_breakdown}: Cross-judge robustness analyses, including GPT-4o and DeepSeek re-judging, source-stratified analysis, and open-weight evaluation.
    \item Appendix~\ref{app:additional_discussion}: Additional discussion (discoverability drivers, error taxonomy, practical directions, precision details).
    \item Appendix~\ref{app:human_baseline}: Human baseline protocol and results.
    \item Appendix~\ref{app:ocr_assisted}: OCR-assisted prompting experiment.
    \item Appendix~\ref{app:eval_details}: Evaluation details (parameters, preprocessing).
\end{itemize}

\section{Full Prompts}
\label{app:prompts}

All conditions share the same image input (base64-encoded, standardized to 1024px maximum long edge preserving aspect ratio) and editing instruction. Differences are the system prompt, user prompt structure, and (for Oracle) additional context.

\subsection{Direct Condition}

\prompttitle{System Prompt}
\begin{promptbox}
You are a visual analysis expert. When given an image and an editing instruction, identify what aspects should be carefully considered to ensure the edit is fully correct and consistent.
\end{promptbox}

\prompttitle{User Prompt}
\begin{promptbox}
Here is an image and an editing instruction.

Editing instruction: {instruction}

Based on the image, what potential issues or considerations should be kept in mind when performing this edit? List everything you think is important.
\end{promptbox}

\paragraph{Design rationale.} No mention of ``implicit constraints,'' ``physical laws,'' or taxonomy terms. The model must discover constraints entirely on its own.

\subsection{Chain-of-Thought (CoT) Condition}

\prompttitle{System Prompt}
\begin{promptbox}
You are a visual analysis expert skilled at systematic, step-by-step reasoning about images. When analyzing an editing task, think through each aspect carefully before reaching your conclusion.
\end{promptbox}

\prompttitle{User Prompt}
\begin{promptbox}
Here is an image and an editing instruction.

Editing instruction: {instruction}

Please analyze step by step:

1. First, identify the exact element being edited and its location in the image.
2. Then, carefully examine the entire image -- are there any other regions or elements that are visually, semantically, or physically related to the editing target?
3. For each related element you find, explain what relationship it has with the editing target and whether it would need to change after the edit.
4. Finally, summarize all the considerations needed for a fully correct and consistent edit.

Think through each step thoroughly before moving to the next.
\end{promptbox}

\paragraph{Design rationale.} Step-by-step reasoning guidance but no specific constraint types. Information parity with \ours{}-Self: both ask the model to look for related elements, but CoT uses natural language steps while \ours{}-Self requires structured JSON.

\subsection{ICG-Self Condition}

\prompttitle{System Prompt}
\begin{promptbox}
You are a visual analysis expert. When given an image and an editing instruction, your task is to construct an Implicit Constraint Graph (ICG) -- a structured representation of all implicit constraints that should be satisfied after the edit.

An implicit constraint is a condition not explicitly stated in the editing instruction but reasonably expected to hold after the edit, arising from visual, semantic, physical, or stylistic dependencies between the edited element and other elements in the image.
\end{promptbox}

\prompttitle{User Prompt}
\begin{promptbox}
Here is an image and an editing instruction.

Editing instruction: {instruction}

Please construct an Implicit Constraint Graph by identifying:

1. Edit target: What exactly is being changed, and where is it in the image?
2. Constraint nodes: What other regions or elements in the image are implicitly affected by this edit? For each one, specify:
   - The affected region
   - The constraint type (physical like reflections/shadows, spatial like mirror/symmetry, semantic like price/label linkage, stylistic like font/color, or cross-modal like scene compatibility)
   - The expected update
   - The reason
3. Inter-constraints: Are there any consistency requirements BETWEEN the affected regions themselves?

Output your analysis as structured JSON:
{
  "edit_target": {
    "element": "...",
    "region": "...",
    "action": "..."
  },
  "constraint_nodes": [
    {
      "id": "n1",
      "region": "...",
      "constraint_type": "...",
      "expected_update": "...",
      "reason": "...",
      "strength": "strong"
    }
  ],
  "inter_constraints": [
    {
      "between": ["n1", "n2"],
      "type": "...",
      "description": "..."
    }
  ]
}

Be thorough but precise -- only include constraints that are genuinely grounded in what you can observe in the image.
\end{promptbox}

\paragraph{Design rationale.} Requires structured JSON graph output. The full 19-subtype taxonomy is \emph{not} provided---only natural language categories. Anti-hallucination instruction included.

\subsection{ICG-Oracle Condition}

\prompttitle{System Prompt}
\begin{promptbox}
You are a visual analysis expert. You will be given an image, an editing instruction, and a pre-constructed Implicit Constraint Graph that describes all the implicit constraints that should be satisfied after the edit. Your task is to analyze the image based on this constraint graph.
\end{promptbox}

\prompttitle{User Prompt}
\begin{promptbox}
Here is an image and an editing instruction.

Editing instruction: {instruction}

A domain expert has identified the following implicit constraints for this edit:

Implicit Constraint Graph:
{oracle_icg_json}

Based on this constraint graph and the image:

1. For each constraint node, verify whether you can identify the described region in the image.
2. For each constraint, explain whether the expected update makes sense given the image.
3. For each inter-constraint, explain whether the consistency requirement is valid.
4. Are there any additional constraints the expert may have missed?

Provide your analysis addressing each constraint node by its ID.
\end{promptbox}

\subsection{Adversarial Oracle Condition}

Identical to ICG-Oracle, except one adversarial constraint node (generated by GPT-4o; see Appendix~\ref{app:adversarial}) is injected alongside the ground-truth nodes. An additional instruction states:

\begin{promptbox}
IMPORTANT: Not all constraints listed above are necessarily valid. For each constraint, evaluate whether it is VALID or INVALID based on what you actually observe in the image.
\end{promptbox}

\subsection{Taxonomy-Only Condition}

The system prompt is identical to ICG-Self. The user prompt provides the full 5-category, 19-subtype taxonomy as a structured checklist and asks the model to check each subtype for applicability---but provides no case-specific constraint nodes.

\prompttitle{User Prompt}
\begin{promptbox}
Here is an image and an editing instruction.

Editing instruction: {instruction}

Below is a taxonomy of implicit constraint types that
can arise in text-in-image editing tasks. For EACH
subtype, examine the image and determine whether this
type of constraint applies to the given edit.

== CONSTRAINT TAXONOMY ==

1. PHYSICAL
   1.1 Reflection: Does the edited text appear on or
       near a reflective surface?
   1.2 Shadow: Does the edited text cast a shadow that
       would need updating?
   1.3 Perspective: Is the text on a 3D or angled
       surface requiring perspective consistency?
   1.4 Occlusion: Is any part of the text partially
       hidden behind another object?

2. SPATIAL
   2.1 Mirror: Does the text appear in a mirrored or
       symmetric arrangement with identical content?
   2.2 Symmetry: Are there symmetric visual elements
       that must remain balanced?
   2.3 Repetition: Does the same text appear multiple
       times in the image?

3. SEMANTIC
   3.1 Numerical: Are there computed or linked numerical
       values (prices, dates, totals)?
   3.2 Categorical: Does a category/brand label require
       associated attributes to match?
   3.3 Functional: Are there structurally bound elements
       (headers with content, legends with charts)?
   3.4 Promotional: Is there promotional logic
       (buy X get Y, limited time) that must stay valid?
   3.5 Visual-Data: Do data visualizations (charts,
       graphs, tables) need to match text labels?

4. STYLE
   4.1 Font: Must other text elements match the
       typography of the edited text?
   4.2 Color: Does a color scheme or palette need to
       remain consistent?
   4.3 Layout: Does the spatial arrangement need to
       accommodate the new content?
   4.4 Material: Does the text appear on a textured
       surface (wood, metal, fabric) requiring
       consistent rendering?

5. CROSS-MODAL
   5.1 Scene: Does the new text conflict with the
       visual scene context?
   5.2 Temporal: Are there time-related elements
       (dates, seasons, clocks) that must stay
       consistent?
   5.3 Cultural: Does the edit introduce cultural,
       linguistic, or tonal mismatches?

For each applicable subtype, describe:
- The specific region affected
- What update is needed
- Why this constraint applies

Only include subtypes that genuinely apply based on
what you observe in the image.
\end{promptbox}

\subsection{NL-Self Condition}
\label{app:nl_self}

NL-Self is a free-form self-discovery condition that asks for the same information dimensions as ICG-Self: region, type, update, and reason. It also changes the task framing, so Appendix~\ref{app:format_ablation} uses a strict format ablation to measure the format effect separately.

\prompttitle{System Prompt}
\begin{promptbox}
You are a visual analysis expert. When given an image and an editing instruction, your task is to identify all implicit constraints that should be satisfied after the edit.

An implicit constraint is a condition not explicitly stated in the editing instruction but reasonably expected to hold after the edit, arising from visual, semantic, physical, or stylistic dependencies between the edited element and other elements in the image.
\end{promptbox}

\prompttitle{User Prompt}
\begin{promptbox}
Here is an image and an editing instruction.

Editing instruction: {instruction}

Please identify all implicit constraints for this edit. For each constraint you find, describe:

1. The affected region in the image
2. The type of constraint (physical, spatial, semantic, stylistic, or cross-modal)
3. What update is needed
4. Why this region is affected by the edit

Be thorough -- identify every region that would need to change as a consequence of the edit, even if the instruction doesn't mention it. Only include constraints that are genuinely grounded in what you can observe in the image.
\end{promptbox}

\paragraph{Design rationale.} NL-Self preserves the four ICG-Self information fields (region, type, update, reason) while changing both format and task framing. The +0.082 recall gap between NL-Self (0.616) and ICG-Self (0.534) therefore combines two effects: structural overhead and dependency-search framing. The strict format ablation (Appendix~\ref{app:format_ablation}) measures the first effect directly: $\Delta = +0.042$ macro recall (+0.038 paired node-level recall, $p < 0.001$).

\subsection{Signal-Guided (SG) Condition}

The Signal-Guided prompt replaces the full taxonomy checklist with a shorter set of concrete search cues. It directs the model to inspect high-risk visual evidence---reflective surfaces, repeated text, numerical relationships, labels, legends, symmetry, style, and scene consistency---and then report any triggered constraints.

\prompttitle{System Prompt}
\begin{promptbox}
You are a visual constraint analyst. Your job is to find implicit constraints that an editing instruction triggers -- regions or elements that MUST change as a consequence of the edit, even though the instruction doesn't mention them.
\end{promptbox}

\prompttitle{User Prompt}
\begin{promptbox}
Here is an image and an editing instruction.

Editing instruction: {instruction}

Your task: Search the image for regions that are IMPLICITLY AFFECTED by this edit -- things that must change even though the instruction doesn't say so.

HIGH PRIORITY signals to check:
- Reflective/mirrored surfaces showing the edited text
- The same text appearing elsewhere in the image
- Numbers that are mathematically derived from the edited value
- Labels, legends, or captions that reference the edited element
- Symmetric or repeated visual structures

ALSO CHECK:
- Whether the edit creates a mismatch with the visual scene
- Whether surrounding text styles must match the edit
- Whether any cultural or temporal context is violated
- Whether 3D perspective or occlusion affects the edit

For each constraint you find, report:
- The specific region affected
- What type of constraint it is
- What update is needed
- Why this region is coupled to the edit

Err on the side of reporting -- if you see something that MIGHT be a coupled region, include it. Only skip constraints that are clearly not applicable.
\end{promptbox}

\paragraph{Design rationale.} Signal-Guided tests whether a compact set of visual search cues can improve discovery without giving case-specific affected regions. Taxonomy-Only enumerates all 19 subtypes; Signal-Guided instead prioritizes cues that commonly reveal secondary dependencies. It also permits free-form output, avoiding the JSON schema used by ICG-Self.

\subsection{Oracle Decomposition Conditions (Region-Only, Reason-Only, Type-Only)}
\label{app:oracle_decomp_prompts}

These three conditions decompose the oracle information into its constituent fields to isolate which component drives the oracle gain. All three share the same image and editing instruction as the main conditions, but provide different subsets of the ground-truth oracle information. They are \emph{not} self-discovery conditions---the number of listed items equals the number of ground-truth constraint nodes in each case (labels may repeat when multiple nodes share a type), so the model knows how many constraints exist.

\subsubsection{Region-Only}

\prompttitle{System Prompt}
\begin{promptbox}
You are a visual analysis expert. You will be given an image, an editing instruction, and a list of image regions that a domain expert has identified as potentially affected by this edit. Your task is to analyze why each region might need to change.
\end{promptbox}

\prompttitle{User Prompt}
\begin{promptbox}
Here is an image and an editing instruction.

Editing instruction: {instruction}

A domain expert has identified the following regions as potentially affected by this edit:

{region_list}

For each region:
1. Explain whether and why it needs to change after the edit.
2. If it needs to change, describe what the update should be.
3. If you think it does NOT need to change, explain why not.
\end{promptbox}

\paragraph{Design rationale.} Provides only spatial localization (which regions are affected), withholding the constraint type and causal reason. Region names are cleaned to remove leakage of type or reason information (e.g., ``the shadow of the text'' $\rightarrow$ ``a region below the main text'').

\subsubsection{Reason-Only}

\prompttitle{System Prompt}
\begin{promptbox}
You are a visual analysis expert. You will be given an image, an editing instruction, and a list of abstract constraint descriptions that a domain expert has identified as relevant to this edit. The descriptions explain WHY a dependency exists but do NOT name the specific affected regions. Your task is to locate the specific regions in the image where each constraint applies.
\end{promptbox}

\prompttitle{User Prompt}
\begin{promptbox}
Here is an image and an editing instruction.

Editing instruction: {instruction}

A domain expert has identified the following types of implicit constraints triggered by this edit. Note: the descriptions explain the causal relationship but do NOT identify the specific affected regions --- you must find them yourself.

{reason_list}

For each constraint:
1. Locate the specific region in the image where this constraint applies.
2. Describe what update is needed for that region.
3. If you cannot find a matching region, say so.
\end{promptbox}

\paragraph{Design rationale.} Provides de-identified causal explanations of why each constraint exists (all region names, object references, and specific text content replaced with generic placeholders such as ``a dependent visual element''), withholding explicit region localization. Models must locate the affected region from the causal logic alone. The de-identification procedure and residual leakage are quantified in Appendix~\ref{app:leakage}.

\subsubsection{Type-Only}

\prompttitle{System Prompt}
\begin{promptbox}
You are a visual analysis expert. You will be given an image, an editing instruction, and a list of constraint type labels that a domain expert has identified as relevant to this edit. Your task is to find where each constraint type applies in the image and what needs to change.
\end{promptbox}

\prompttitle{User Prompt}
\begin{promptbox}
Here is an image and an editing instruction.

Editing instruction: {instruction}

A domain expert has listed one constraint-type label for each ground-truth constraint node below. Labels may repeat, and the number of listed items equals the number of ground-truth constraints in this case.

{type_list}

For each listed constraint type:
1. Find the specific region in the image where this constraint applies.
2. Explain why this constraint type is relevant here.
3. Describe what update is needed.
4. If you cannot identify where this constraint applies, say so.
\end{promptbox}

\paragraph{Design rationale.} Provides only the constraint type label (e.g., \texttt{Semantic\_Numerical}), withholding both the specific region and the causal reason. Unlike Taxonomy-Only---which provides all 19 subtypes as a checklist for self-discovery---Type-Only provides oracle-confirmed type labels for the $N$ constraints that actually exist in the case, isolating the contribution of constraint-type information to the oracle gain.

\section{Automated Judge Prompt}
\label{app:judge}

The automated judge (Claude Sonnet 4.6) evaluates each ground-truth constraint independently. Temperature is set to 0.

\begin{promptbox}
You are a strict academic evaluator assessing whether an AI model's response genuinely identified a specific implicit constraint.

== EDITING INSTRUCTION ==
{instruction}

== CONSTRAINT TO CHECK ==
Region: {constraint.region}
Constraint type: {constraint.constraint_type}
Expected update: {constraint.expected_update}
Reason: {constraint.reason}

== MODEL RESPONSE ==
{model_response[:2000]}

== YOUR TASK ==
Did the model response GENUINELY identify this specific constraint?

STRICT criteria -- answer YES only if the model:
1. Specifically named or described the affected region
2. Explicitly stated that it needs to be updated/changed
3. Gave a reason that matches the actual constraint

Answer NO if the model:
- Only gave generic editing advice without addressing this specific logical dependency
- Mentioned the region but didn't say it needs to change
- Got the right region but wrong reason

Respond ONLY in JSON format:
{"found": true/false, "confidence": "high"/"medium"/"low", "reasoning": "one sentence"}
\end{promptbox}

\section{Complete Constraint Taxonomy}
\label{app:taxonomy_full}

\subsection{Taxonomy Development}

Our taxonomy development followed a rigorous four-phase pipeline to ensure theoretical grounding and empirical validity:
\begin{enumerate}
    \item \textbf{Phase 1: Seed from Literature.} We grounded the high-level constraint categories in scene graph relationship modeling~\citep{krishna2017visual}, controllable generation structural conditions~\citep{zhang2023adding}, and cognitive frameworks of feature binding and perceptual organization~\citep{treisman1980feature,treisman1982illusory,greff2020binding}.
    \item \textbf{Phase 2: Open Coding and Extraction.} We performed open coding on 300+ knowledge-prompt annotated cases from \teb{}~\citep{gui2025texteditbench} to extract recurring edit-induced constraint patterns.
    \item \textbf{Phase 3: Pilot Annotation and Consolidation.} We conducted two rounds of pilot annotation (10 cases each, yielding 60 constraint nodes in total) with three annotators to resolve boundaries and merge categories. Subtypes were treated as stabilized once no new type appeared in five consecutive cases during iterative pilot expansion. For instance, a tentative top-level ``Functional'' category was folded into the broader Semantic category, with Functional retained as a subtype. Subtype-level classification reached substantial agreement ($\kappa = 0.77$).
    \item \textbf{Phase 4: Taxonomy Freezing.} The taxonomy of 19 subtypes was frozen \emph{before} curating non-\teb{} cases from Pexels, Unsplash, and Canva. This separates taxonomy construction from the evaluation sources, while acknowledging that Phase~2 used \teb{} pilot cases.
\end{enumerate}

\subsection{Disambiguation Guidelines}

Several subtype pairs share surface similarities.
To ensure consistent annotation, we established the following decision rules (annotators received these as part of their guidelines):

\paragraph{Reflection vs.\ Mirror.}
\texttt{physical\_reflection} applies when the constraint concerns whether a reflection \emph{physically exists and looks correct} (e.g., water surface should show a flipped image of the text).
\texttt{spatial\_mirror} applies when a mirror already shows text and the \emph{content} must be updated to match the edited source.
Decision rule: if the edit requires \emph{creating or correcting the physics} of a reflection, use Reflection; if it requires \emph{synchronizing content} across a real object and its mirror image, use Mirror.

\paragraph{Categorical vs.\ Functional.}
\texttt{semantic\_categorical} applies when changing a \emph{class, type, or descriptive label} requires updating associated descriptive attributes (e.g., changing an airline name requires updating the logo).
\texttt{semantic\_functional} applies when changing text that serves a \emph{structural or interactive role} requires updating dependent references (e.g., swapping two sections in a document requires headers to follow their content).
Decision rule: if the dependency is driven by \emph{what the element describes}, use Categorical; if by \emph{what the element does in a structure}, use Functional.

\paragraph{Numerical vs.\ Promotional.}
\texttt{semantic\_numerical} applies when the dependency is \emph{mathematically computable} (e.g., discount 30\% on \$100 $\rightarrow$ sale price must be \$70).
\texttt{semantic\_promotional} applies when the dependency is \emph{messaging coherence} that is not a calculation (e.g., ``HALF PRICE'' label is no longer accurate after changing the discount to 30\%).
A single edit can trigger both types simultaneously; they are not mutually exclusive at the case level, though each constraint \emph{node} receives exactly one label.

\paragraph{Scene vs.\ Temporal vs.\ Cultural.}
All three are cross-modal coherence subtypes.
\texttt{xmod\_temporal} applies when the text contains an \emph{explicit time reference} (date, hour, season name) that conflicts with visual temporal cues.
\texttt{xmod\_cultural} applies when the text's \emph{register, formality, or cultural convention} mismatches the context (e.g., slang on a government form).
\texttt{xmod\_scene} applies to all other text--scene incompatibilities (e.g., wrong language for the geographic locale).
Decision rule: explicit time $\rightarrow$ Temporal; tone/register $\rightarrow$ Cultural; everything else $\rightarrow$ Scene.

\subsection{Ground-Truth Existence Re-Verification}
\label{app:gt_reverify}

The Fleiss' $\kappa = 0.77$ reported in the main paper measures agreement on \emph{subtype classification}, i.e.\ which of the 19 subtypes a given constraint belongs to. It does not, by itself, establish that each annotated constraint genuinely exists, that its affected region and expected update are correct, or that the provided dependency reason is visually grounded and valid. To address this directly, we re-verify a $20\%$ stratified sample of all 776 ground-truth nodes (156 nodes, allocated proportionally across the five categories: physical 34, spatial 11, semantic 33, style 35, cross-modal 43).

\paragraph{Protocol.} An independent verification annotator, blind to the original annotation rationale, is shown only the source image, the editing instruction, and the four claimed fields of each node (region, type, expected update, reason). For each node, the annotator records a binary judgment on four dimensions: (D1) does the constraint genuinely exist, i.e.\ does the edit actually require this region to change; (D2) is the claimed affected region correct; (D3) is the expected update correct and necessary; and (D4) is the provided dependency reason visually grounded and logically sufficient to explain why this region should be updated under the instruction. Because the original ground truth asserts ``yes'' on all dimensions by construction, the per-dimension fraction of ``yes'' responses equals the agreement-with-ground-truth rate. Per-dimension agreement rates are shown in Table~\ref{tab:gt_reverify}.

\begin{table}[h]
\centering
\small
\caption{Ground-truth existence re-verification on a 20\% stratified sample (156 nodes). Agreement-with-ground-truth per dimension.}
\label{tab:gt_reverify}
\setlength{\tabcolsep}{5pt}
\begin{tabular}{@{}lc@{}} 
\toprule 
\textbf{Dimension} & \textbf{Agree w/ GT} \\ 
\midrule 
D1 Existence & 90.4\% \\ 
D2 Region correct & 84.6\% \\ 
D3 Update correct & 89.1\% \\ 
D4 Reason valid & 86.5\% \\ 
\bottomrule 
\end{tabular}
\end{table}

\noindent Agreement rates are computed over 156 sampled nodes verified by an independent annotator blind to the original rationale.
D1 existence agreement (90.4\%) indicates that the vast majority of annotated constraints are judged to genuinely require a change after the edit. Lower agreement on D2 region (84.6\%), D3 update (89.1\%), and D4 reason validity (86.5\%) reflects the inherent subjectivity in specifying exact spatial boundaries, necessary update descriptions, and visually grounded dependency rationales, consistent with the ``clearly implied vs.\ optional'' boundary noted in the main paper.

\subsection{Full Taxonomy Table}

Table~\ref{tab:taxonomy_full} lists all 5 categories and 19 subtypes with descriptions, examples, and node counts.

\begin{table*}[t]
\centering
\small
\caption{Complete constraint taxonomy: 5 categories, 19 subtypes, with descriptions and examples. $N$ = constraint nodes across 461 cases.}
\label{tab:taxonomy_full}
\begin{tabular}{@{}llp{5cm}p{5cm}r@{}}
\toprule
\textbf{Category} & \textbf{Subtype} & \textbf{Description} & \textbf{Example} & \textbf{$N$} \\
\midrule
\multirow{4}{*}{Physical}
  & Reflection & Reflected object must update & Changing storefront text; window reflection must match & 50 \\
  & Shadow & Shadow must match new content & Modifying raised letters; shadow shape must update & 29 \\
  & Perspective & 3D text consistency & Changing text on angled sign; perspective must match & 20 \\
  & Occlusion & Partially hidden text consistency & Editing text behind an object; hidden portion must cohere & 69 \\
\midrule
\multirow{3}{*}{Spatial}
  & Mirror & Mirrored text must both update & Symmetric sign with mirrored text & 14 \\
  & Symmetry & Symmetric elements must remain so & Editing one of two symmetric labels & 18 \\
  & Repetition & Repeated instances must all update & Brand name appearing multiple times & 24 \\
\midrule
\multirow{5}{*}{Semantic}
  & Numerical & Linked values must remain consistent & Changing discount; derived prices must update & 19 \\
  & Categorical & Category labels must match content & Changing airline name; logo must match & 54 \\
  & Functional & Bound elements must move together & Swapping sections; headers stay with content & 21 \\
  & Promotional & Promotional logic must remain valid & Changing ``Buy 2 Get 1 Free'' & 20 \\
  & Visual-Data & Visualizations must match labels & Changing a chart title; data must align & 52 \\
\midrule
\multirow{4}{*}{Style}
  & Font & Typography must remain consistent & Editing one heading; others should match font & 67 \\
  & Color & Color scheme must remain harmonious & Changing team name; jersey colors should match & 29 \\
  & Layout & Arrangement must accommodate new content & Longer text; surrounding layout must adjust & 21 \\
  & Material & Material rendering must match surface & Text on wood; new text must appear carved & 56 \\
\midrule
\multirow{3}{*}{Cross-modal}
  & Scene & Scene semantics must remain consistent & Changing store name; context must be compatible & 75 \\
  & Temporal & Time-related elements must be consistent & Changing date; day-of-week must match & 36 \\
  & Cultural & Cultural conventions must be respected & Changing text to different language & 102 \\
\midrule
\multicolumn{4}{l}{\textbf{Total}} & \textbf{776} \\
\bottomrule
\end{tabular}
\end{table*}

\section{Full Per-Subtype Results}
\label{app:subtype_full}

Table~\ref{tab:subtype_full} reports Direct and Oracle Constraint Recall for all 19 subtypes, sorted by Direct recall.

\begin{table*}[t]
\centering
\small
\caption{Full per-subtype Constraint Recall (micro-averaged, 4 models) under Direct and Oracle conditions. $N$ = number of constraint nodes across 461 cases. All subtypes have $N \geq 14$. Sorted by Direct recall.}
\label{tab:subtype_full}
\begin{tabular}{@{}llcccr@{}}
\toprule
\textbf{Category} & \textbf{Subtype} & \textbf{Direct} & \textbf{Oracle} & \textbf{$\Delta$} & \textbf{$N$} \\
\midrule
Cross-modal & Scene & 0.137 & 0.940 & +0.803 & 75 \\
Cross-modal & Cultural & 0.179 & 0.917 & +0.738 & 102 \\
Spatial & Symmetry & 0.208 & 0.958 & +0.750 & 18 \\
Semantic & Functional & 0.274 & 0.857 & +0.583 & 21 \\
Physical & Occlusion & 0.348 & 0.935 & +0.587 & 69 \\
Style & Layout & 0.381 & 0.964 & +0.583 & 21 \\
Style & Color & 0.405 & 0.784 & +0.379 & 29 \\
Semantic & Categorical & 0.454 & 0.884 & +0.431 & 54 \\
Cross-modal & Temporal & 0.458 & 0.931 & +0.472 & 36 \\
Semantic & Promotional & 0.463 & 0.938 & +0.475 & 20 \\
Spatial & Repetition & 0.542 & 0.938 & +0.396 & 24 \\
Style & Font & 0.552 & 0.862 & +0.310 & 67 \\
Physical & Shadow & 0.586 & 1.000 & +0.414 & 29 \\
Physical & Perspective & 0.588 & 0.988 & +0.400 & 20 \\
Spatial & Mirror & 0.625 & 0.982 & +0.357 & 14 \\
Style & Material & 0.679 & 0.929 & +0.250 & 56 \\
Physical & Reflection & 0.690 & 0.970 & +0.280 & 50 \\
Semantic & Visual-Data & 0.731 & 0.889 & +0.159 & 52 \\
Semantic & Numerical & 0.868 & 0.987 & +0.118 & 19 \\
\midrule
\multicolumn{2}{l}{\textbf{Overall}} & 0.447 & 0.921 & +0.475 & 776 \\
\bottomrule
\end{tabular}
\end{table*}

\section{$E_\text{inter}$ Ablation Details}
\label{app:einter}

\begin{table}[t]
\centering
\small
\caption{$E_\text{inter}$ ablation: Constraint Recall under Full Oracle vs.\ Oracle without inter-constraint edges (228 cases with $E_\text{inter}$, 912 evaluations). Deltas computed from unrounded values.}
\label{tab:einter_full}
\begin{tabular}{@{}lccc@{}}
\toprule
\textbf{Model} & \textbf{Full} & \textbf{NoInter} & \textbf{$\Delta$} \\
\midrule
Claude Opus 4.6 & 0.872 & 0.817 & $+$0.055 \\
GPT-4o & 0.978 & 0.977 & $+$0.001 \\
Gemini 2.5 Flash & 0.896 & 0.920 & $-$0.023 \\
Qwen3-VL-Plus & 0.907 & 0.915 & $-$0.007 \\
\midrule
\textbf{All Models} & \textbf{0.913} & \textbf{0.907} & $+$\textbf{0.007} \\
\bottomrule
\end{tabular}
\end{table}

Table~\ref{tab:einter_full} shows a small, directionally mixed effect. Claude and GPT-4o improve slightly with the full oracle, while Gemini and Qwen3 improve slightly after removing $E_\text{inter}$. The pooled difference is $\Delta = +0.007$, so the oracle gain is driven primarily by node identity and dependency reason---which region is affected and why---rather than by optional inter-node consistency edges.

\section{Oracle Recall by Category and Model}
\label{app:oracle_model}

\begin{table}[t]
\centering
\small
\caption{Oracle Recall by category and model (461 cases), micro-averaged over constraint nodes within each category. Per-model overall values differ from the macro case-level Oracle recall reported in Table~1 of the main paper because of the different aggregation method.}
\label{tab:oracle_model}
\begin{tabular}{@{}lcccc@{}}
\toprule
\textbf{Category} & \textbf{Opus} & \textbf{GPT-4o} & \textbf{Gemini} & \textbf{Qwen3} \\
\midrule
Physical & 0.946 & 0.982 & 0.958 & 0.964 \\
Spatial & 0.929 & 1.000 & 0.946 & 0.946 \\
Semantic & 0.928 & 0.952 & 0.819 & 0.904 \\
Style & 0.827 & 0.960 & 0.867 & 0.879 \\
Cross-modal & 0.887 & 0.977 & 0.934 & 0.911 \\
\midrule
\textbf{Overall} & 0.898 & 0.970 & 0.901 & 0.916 \\
\bottomrule
\end{tabular}
\end{table}

Table~\ref{tab:oracle_model} breaks down Oracle recall by category and model. GPT-4o is the strongest at \emph{using} provided constraints (0.970 micro-averaged Oracle recall in this category breakdown). Style constraints show the lowest Oracle recall for Claude (0.827), while Gemini's lowest Oracle recall is Semantic (0.819) and its Style recall is also relatively low (0.867). This pattern suggests that stylistic and semantic constraint interpretation can remain difficult even with oracle guidance. In non-oracle conditions, Claude is the strongest at \emph{discovering} constraints (Direct: 0.659), while GPT-4o shows the strongest CoT improvement (+0.166) but the weakest Direct performance (0.320).

\section{Oracle Decomposition}
\label{app:oracle_decomp}

To isolate which oracle components drive the performance boost, we run three ablation conditions on all 461 cases $\times$ 4 models. All results in this section use macro-averaged case-level recall (overall Full Oracle: 0.938) to align with Table~1 of the main text. Full prompts for all three conditions appear in Appendix~\ref{app:oracle_decomp_prompts}. The ablation conditions are:
\begin{itemize}
    \item \textbf{Region-Only}: Provides only which regions are affected (spatial localization), without explaining why or what constraint type applies.
    \item \textbf{Reason-Only}: Provides abstract causal explanations of why each constraint exists, with all region names, object references, and specific text content systematically removed (replaced with generic placeholders such as ``a dependent visual element'' or ``a related text-bearing region''). Models must locate affected regions independently based on the causal logic alone.
    \item \textbf{Type-Only}: Provides only the constraint type label (e.g., ``Semantic\_Numerical''), without naming the region or explaining why.
\end{itemize}

The Reason-Only descriptions are de-identified causal explanations. GPT-4o rewrites each reason to preserve the abstract causal relation while removing explicit region names, object references, and verbatim text content, replacing them with placeholders such as ``a dependent visual element.'' We manually reviewed all 776 rewritten descriptions for lexical leakage and found no remaining region names or verbatim text. De-identification reduces direct localization cues; it does not remove semantic implication. For example, a reason about a displayed total derived from a per-unit price still implies that a price-like region exists. Appendix~\ref{app:leakage} quantifies this residual signal. Table~\ref{tab:oracle_decomp_full} reports the decomposition results.

\begin{table}[t]
\centering
\small
\caption{Oracle decomposition results (461 cases $\times$ 4 models). Reason-Only provides de-identified causal explanations; models must independently locate affected regions.}
\label{tab:oracle_decomp_full}
\begin{tabular}{@{}lcccc@{}}
\toprule
\textbf{Model} & \textbf{Region} & \textbf{Type} & \textbf{Reason} & \textbf{Full Oracle} \\
\midrule
Claude Opus 4.6 & 0.739 & 0.848 & \textbf{0.923} & 0.916 \\
GPT-4o & 0.511 & 0.547 & 0.775 & 0.976 \\
Gemini 2.5 Flash & 0.645 & 0.639 & 0.723 & 0.931 \\
Qwen3-VL-Plus & 0.544 & 0.550 & 0.705 & 0.927 \\
\midrule
\textbf{All} & 0.610 & 0.646 & \textbf{0.782} & 0.938 \\
\bottomrule
\end{tabular}
\end{table}

\textbf{Key findings:}
\begin{enumerate}
    \item \textbf{Reason-Only} (0.782) substantially outperforms both Region-Only (0.610, $\Delta = +0.172$) and Type-Only (0.646, $\Delta = +0.136$). Abstract causal explanations provide stronger usable cues for locating affected regions than explicit region names alone, while still retaining semantic and partial-localization information (quantified below).
    \item \textbf{Type-Only} (0.646) and \textbf{Region-Only} (0.610) trail Reason-Only by substantial margins. Constraint type labels help more than spatial localization alone in this cleaned run, but neither is sufficient without understanding the causal dependency.
    \item \textbf{Claude under Reason-Only}: Claude achieves higher recall under Reason-Only (0.923) than under Full Oracle (0.916), consistent with its strong performance on abstract causal descriptions and its lower tolerance for information-dense oracle prompts.
    \item The gap between Reason-Only (0.782) and Full Oracle (0.938) represents constraints where causal reasoning alone is insufficient---models additionally need explicit region localization or expected update information to succeed.
\end{enumerate}

Reason-Only provides the strongest partial oracle signal: causal descriptions with explicit region names removed recover 0.782 recall. The result should be read as evidence for the value of causal dependency information, not as a claim that reasons contain no localization signal. The leakage analysis below shows that de-identification removes most exact region cues while preserving the abstract causal category.

\subsection{Quantifying Residual Localization Leakage in De-identified Reasons}
\label{app:leakage}

Reason-Only intentionally removes explicit region names, but causal descriptions can still imply where to look. We measure this residual localization signal with an information leakage test.

\textbf{Setup.} For each of the 776 constraint nodes we give an out-of-pool probe model---DeepSeek-V4-flash (neither evaluated nor used as the de-identifier; GPT-4o performed de-identification)---\emph{only} the reason text, with no image, no editing instruction, and no region information. The probe model attempts two recovery tasks: (i) \textbf{subtype recovery}, a 19-way forced-choice classification graded by exact match (chance $= 1/19 = 0.053$; majority class $= 0.130$); and (ii) \textbf{region recovery}, a free-text guess of the affected region, graded as \texttt{match} / \texttt{partial} / \texttt{none} against the ground-truth region by a separate DeepSeek-V4-flash judge call. We run two arms: \textbf{anon} (the de-identified \texttt{reason\_anon} used in the paper) and \textbf{orig} (the raw reason, a positive-control ceiling that still contains region names, text content, and numbers). The orig$-$anon difference isolates how much de-identification actually removed.

\begin{table}[t]
\centering
\small
\caption{Information leakage test: recovery from reason text alone (out-of-pool DeepSeek probe model; no image, instruction, or region shown). \emph{anon} is the de-identified reason used in the paper; \emph{orig} is the raw reason (positive-control ceiling). Region recovery is judged against the ground-truth region. The orig$-$anon gap measures information removed by de-identification.}
\label{tab:leakage}
\begin{tabular}{@{}lccc@{}}
\toprule
 & \textbf{Subtype} & \textbf{Region} & \textbf{Region} \\
\textbf{Arm} & \textbf{(19-way)} & \textbf{exact} & \textbf{exact+partial} \\
\midrule
anon (de-identified) & 0.680 & 0.190 & 0.672 \\
orig (raw ceiling)   & 0.742 & 0.522 & 0.848 \\
\midrule
De-id removed ($\Delta$) & $-0.062$ & $-0.332$ & $-0.176$ \\
\midrule
\emph{chance / majority} & 0.053 / 0.130 & --- & --- \\
\bottomrule
\end{tabular}
\end{table}

\textbf{Findings.} Table~\ref{tab:leakage} reports the recovery rates. De-identification removes most exact localization signal: region recovery falls from 0.522 (raw) to 0.190 (de-identified), a 64\% relative reduction, and unusable region guesses rise from 0.152 to 0.328. Reason-Only is therefore not simply telling the model where to look. Residual signal remains: from a de-identified reason alone, the probe recovers the exact region 19.0\% of the time and the constraint subtype 68.0\% of the time, far above the 5.3\% chance and 13.0\% majority-class baselines. Subtype recovery drops only modestly after de-identification ($\Delta = -0.062$), because the causal category is intrinsic to the reason; a reason describing a derived quantity implies a numerical constraint regardless of which region names are stripped.

The leakage pattern varies by constraint category. In the anon arm, physical (region-exact 0.38) and spatial (0.36) reasons leak the most location signal because reflections, shadows, and mirrored elements have canonical spatial relationships to the edited text; cross-modal reasons leak the least (0.08). De-identification removes the majority of pinpoint localization while preserving the abstract causal category that drives Reason-Only performance. The Reason-Only advantage over Region-Only ($+0.172$) cannot be explained by residual localization alone: the de-identified reasons retain only one-third of the raw exact-localization signal yet still outperform explicit region guidance.

\section{Failure Source Decomposition}
\label{app:failure}

We decompose failures into three categories:
\begin{itemize}
    \item \textbf{Discovery Failure}: Constraint missed under non-oracle conditions but found under Oracle.
    \item \textbf{Use Failure}: Constraint provided in Oracle but model still fails---mentions the constraint but reaches the wrong conclusion.
    \item \textbf{Grounding Failure}: Model understands the constraint but cannot visually locate the relevant region.
\end{itemize}

Macro case-level Oracle recall of 0.938 corresponds to a residual failure rate of about 6.2\%; at the node-level micro granularity (0.921), the residual is 7.9\%. GPT-4o achieves 0.976 macro Oracle recall (2.4\% macro residual), while Claude has 0.916 (8.4\% macro residual), concentrated in style constraints.

Seven subtypes achieve near-perfect Oracle recall ($\geq 0.95$ averaged across all models): Physical\_Shadow (1.000), Physical\_Perspective (0.988), Semantic\_Numerical (0.987), Spatial\_Mirror (0.982), Physical\_Reflection (0.970), Style\_Layout (0.964), and Spatial\_Symmetry (0.958).
For these subtypes, most errors occur before the constraint is surfaced: once provided, models usually apply it correctly.

\section{Adversarial Constraint Generation}
\label{app:adversarial}

For each case, we generate one adversarial constraint using GPT-4o to ensure independence from the Claude judge family. The generation prompt:

\begin{promptbox}
You are helping design an adversarial control experiment.

Given an image and editing instruction, create ONE plausible but factually INCORRECT implicit constraint. The constraint should:
1. Use the same format as real constraints
2. Sound reasonable without seeing the image
3. Be clearly wrong upon inspecting the image
4. Use a constraint_type from the taxonomy

Editing instruction: {instruction}

Output ONE adversarial constraint in JSON format.
\end{promptbox}

\paragraph{Quality control.}
A human annotator reviewed a stratified sample of 54 adversarial constraints:
\begin{itemize}
    \item \textbf{Well-designed} (47/54, 87\%): Surface-plausible but clearly incorrect upon image inspection.
    \item \textbf{Borderline} (4/54, 7\%): Model acceptance is defensible.
    \item \textbf{Too easy} (3/54, 6\%): Obviously wrong without careful inspection.
\end{itemize}
On the 47 well-designed items, the rejection rate is 91.5\% (43/47).

On the full 461-case set (1{,}844 evaluations = 461 cases $\times$ 4 models), the overall rejection rate is 90.5\% (1668/1844). Per-model: Claude 95.0\%, Gemini 92.0\%, Qwen3 90.7\%, GPT-4o 84.2\%.

\paragraph{Informative exceptions.}
Across all 1844 evaluations, models accept 176 adversarial constraints. Accepted cases cluster around borderline-plausible constraints that are partially defensible from the image content. The control therefore tests more than blanket rejection while still showing that models reject most invalid oracle content.

\section{Format Ablation: Strict Minimal Pair}
\label{app:format_ablation}

To isolate the effect of output format on constraint discovery from potential confounds in the NL-Self vs.\ ICG-Self comparison (see the main paper, Results), we conduct a strict format ablation with minimal pair prompts.

\subsection{Experimental Design}

All three conditions share \emph{exactly} the same system prompt and task description (including information dimensions: region, type, update, reason). The \emph{only} difference is the output format instruction appended to the user prompt:

\begin{itemize}
    \item \textbf{Format-JSON}: ``Output your analysis as a JSON array: [\{region, constraint\_type, expected\_update, reason\}]''
    \item \textbf{Format-Markdown}: ``Output your analysis as a Markdown table: $|$ Region $|$ Constraint Type $|$ Expected Update $|$ Reason $|$''
    \item \textbf{Format-FreeForm}: ``Describe each constraint clearly, making sure to cover all four aspects (region, type, update, reason) for each one.''
\end{itemize}

This design eliminates the framing differences between NL-Self and ICG-Self (different system prompts, anti-hallucination instructions, JSON schema) and isolates the pure format effect.

\subsection{Results}

\begin{table}[t]
\centering
\small
\caption{Strict format ablation: Constraint Recall (macro-averaged, 461 cases) under three output formats with identical task prompts. $\Delta$ = FreeForm minus JSON.}
\label{tab:format_ablation}
\begin{tabular}{@{}lcccc@{}}
\toprule
\textbf{Model} & \textbf{JSON} & \textbf{Markdown} & \textbf{FreeForm} & \textbf{$\Delta$} \\
\midrule
Claude Opus 4.6 & 0.706 & 0.754 & 0.717 & +0.010 \\
GPT-4o & 0.411 & 0.430 & 0.499 & +0.087 \\
Gemini 2.5 Flash & 0.549 & 0.584 & 0.566 & +0.017 \\
Qwen3-VL-Plus & 0.502 & 0.582 & 0.553 & +0.052 \\
\midrule
\textbf{All} & \textbf{0.542} & \textbf{0.588} & \textbf{0.584} & \textbf{+0.042} \\
\bottomrule
\end{tabular}
\end{table}

\textbf{Key observations:}
\begin{enumerate}
    \item Free-form output exceeds JSON under the matched-prompt comparison: $\Delta = +0.038$ micro-averaged node-level ($n=3{,}089$ paired decisions, $p < 0.001$); macro-averaged case-level $\Delta = +0.042$. All four models show positive deltas.
    \item The effect is model-dependent. GPT-4o shows the largest format penalty ($\Delta = +0.087$, macro); Claude shows the smallest ($\Delta = +0.010$).
    \item Markdown performs similarly to FreeForm overall (0.588 vs.\ 0.584), so the observed format cost appears concentrated in rigid JSON rather than lightweight tabular structure. The ordering is model-dependent: Claude, Gemini, and Qwen3 perform best under Markdown, while GPT-4o performs best under FreeForm.
    \item The matched format effect accounts for roughly half of the NL-Self vs.\ ICG-Self gap in the main experiment ($+0.042$ vs.\ $+0.082$); the remaining half is consistent with differences in task framing.
\end{enumerate}

Overall, the matched comparison shows that rigid JSON formatting imposes a measurable cost, while lightweight Markdown is much less costly.

\section{Qualitative Case Studies}
\label{app:case_examples}

\begin{figure*}[t]
\centering
\includegraphics[width=0.95\textwidth]{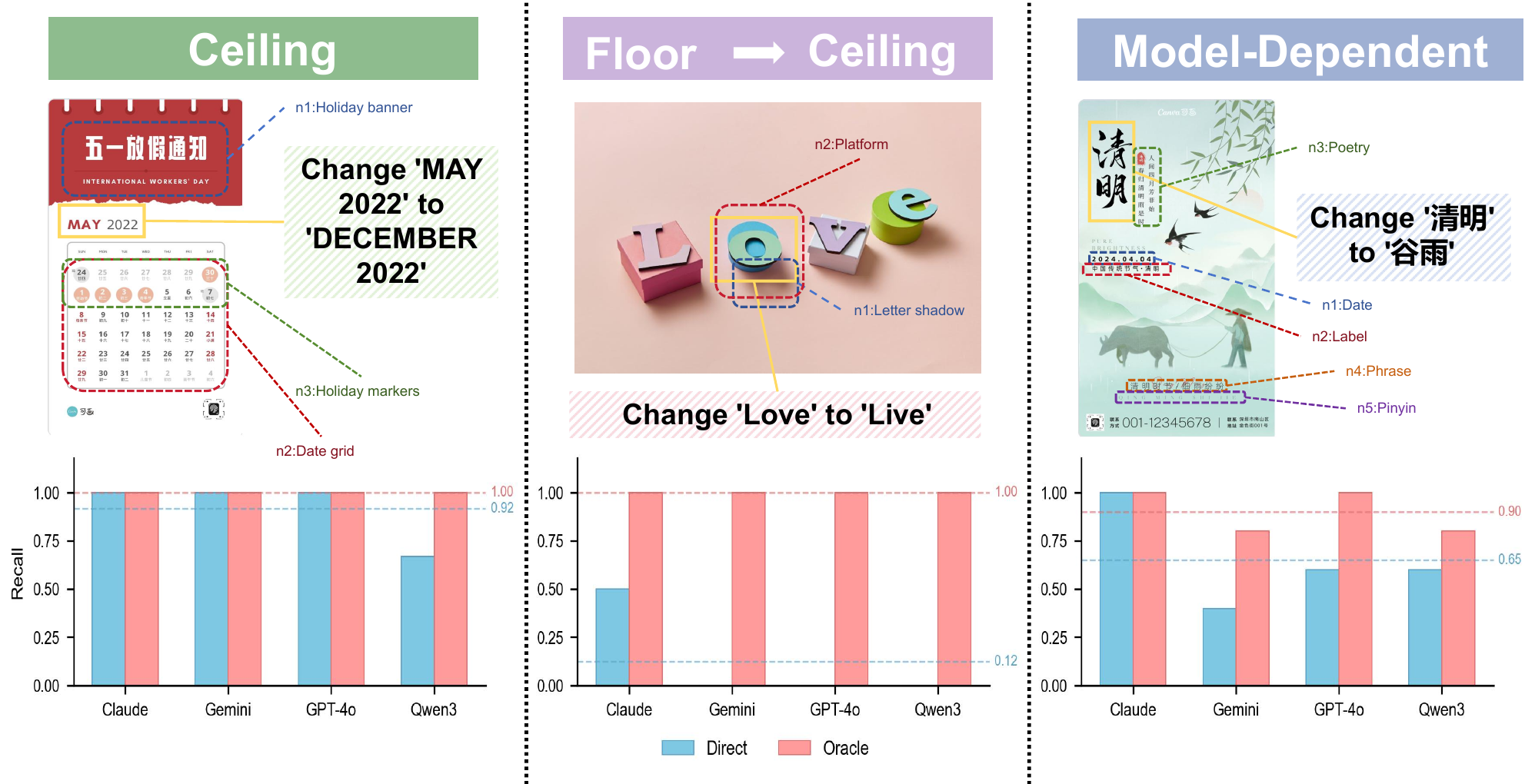}
\caption{Three cases illustrating the discoverability spectrum. \textbf{Left}: Calendar month change from May to December (directly discoverable---most models identify that holidays, date grids, and banner themes must update; Direct recall 0.92). \textbf{Center}: Word edit `Love' $\rightarrow$ `Live' (replacing the letter `o' with `i'; oracle-activated---the cast shadow must change from a round `o' silhouette to a narrow `i' silhouette, but almost no model discovers this without guidance; Direct 0.12 $\rightarrow$ Oracle 1.00). \textbf{Right}: Solar term change from Qingming to Grain Rain (model-dependent---five dependent constraints spanning categorical and temporal subtypes, including date, label, poetry, phrase, and pinyin, must all update, but discovery varies from 2/5 to 5/5 across models).}
\label{fig:case_studies_supp}
\end{figure*}

Figure~\ref{fig:case_studies_supp} presents three cases that span the discoverability spectrum---from directly discoverable to oracle-activated to model-dependent---which we analyze in detail below.

\paragraph{Case 448 (Ceiling): Calendar month change.}
The instruction ``Change month from MAY to DECEMBER'' on a Chinese calendar triggers three Cross-modal\_Temporal constraints: (1) the holiday-themed banner must change from a May Day (Labor Day) theme to a New Year or winter theme, (2) the date grid must reflect December's day-of-week alignment and lunar calendar correspondences, and (3) May-specific holidays (Labor Day, Youth Day, Mother's Day) must be removed or replaced with December events.
Three of four models achieve near-perfect recall under Direct (avg 0.92); all four achieve perfect recall under Oracle.
Temporal consistency between month labels and associated cultural/calendrical content is a highly discoverable constraint---models spontaneously recognize these dependencies without scaffolding.

\paragraph{Case 305 (Oracle-Activated): Letter-edit shadow (`Love'$\rightarrow$`Live').}
The instruction ``Replace the letter `o' with an `i'\,'' (editing the word `Love' to `Live') on a set of 3D letters mounted on a surface triggers two Physical Shadow constraints: (1) the shadow cast by the round `o' must change to a narrow, tall silhouette matching the `i' character, and (2) the shadow of the circular supporting disc must update if the platform shape changes.
Under Direct, recall averages only 0.12---models focus on the letter replacement itself without considering the physical consequence for cast shadows.
Under Oracle, \emph{all four models score 100\%}.
This case shows a low-signal physical dependency: models rarely inspect the shadow under Direct prompting, but recover the constraint once the affected region is specified.

\paragraph{Case 313 (Model-Dependent): Solar term categorical cascade.}
The instruction ``Change `Qingming' to `Grain Rain' '' on a traditional Chinese solar term poster triggers five dependent constraints spanning Semantic\_Categorical and Cross-modal\_Temporal subtypes: 
(1) the date ``2024.04.04'' must change to ``2024.04.19'' (the date of Grain Rain in 2024; Cross-modal\_Temporal), 
(2) the descriptive label ``Chinese traditional solar term $\cdot$ Qingming'' must update to reference Grain Rain (Semantic\_Categorical), 
(3) the vertical poetry referencing Qingming themes must change to Grain Rain-appropriate content (Semantic\_Categorical), 
(4) the descriptive phrase ``Qingming season / light rain remembrance'' must be replaced with a Grain Rain characterization (Semantic\_Categorical), and 
(5) the pinyin ``QING MING SHI JIE'' must become ``GU YU SHI JIE'' (Semantic\_Categorical).
Claude discovers all five constraints under Direct (recall 1.0); GPT-4o and Qwen3 each find three (0.6); Gemini finds only two (0.4).
Under Oracle, all models reach 0.8--1.0.
This case reveals how categorical knowledge depth varies across models: all recognize the date must change, but only Claude spontaneously identifies the poetry and pinyin as dependent elements.

\paragraph{Cross-case patterns.}
Temporal constraints with explicit month--holiday mappings (Case~448) are highly discoverable and require no scaffolding.
Physical shadow constraints (Case~305) are rarely discovered under Direct prompting but universally recoverable under Oracle: the physical consequence is visible, but models usually miss it without explicit prompting.
Categorical cascades (Case~313) expose model-specific variation: the same semantic dependency chain is fully traced by one model but only partially traced by others, reflecting differences in cultural and domain knowledge activation rather than reasoning capacity per se.
The three cases illustrate different discovery regimes: direct temporal dependencies, low-signal physical dependencies, and knowledge-dependent categorical cascades.

\section{Statistical Robustness Analysis}
\label{app:statistical_robustness}

To address the nested structure of our data (constraints within cases, cases evaluated by multiple models), we supplement the main-text McNemar tests and bootstrap intervals with two additional analyses:

\subsection{GEE Logistic Regression}

We fit a generalized estimating equation (GEE) model with binomial family, logit link, and exchangeable working correlation structure, clustering on case ID. Fixed effects: prompting condition (reference: Direct) and model (reference: GPT-4o). Robust standard errors account for within-cluster dependence.

\begin{table}[h]
\centering
\small
\setlength{\tabcolsep}{4pt} 
\caption{GEE logistic regression: fixed effects on constraint discovery (found = 1). $N = 12{,}416$ constraint-level observations across 461 cases.}
\begin{tabular}{@{}lrrrr@{}}
\toprule
\textbf{Effect} & \textbf{Coef.} & \textbf{OR} & \textbf{95\% CI} & \textbf{$p$} \\
\midrule
Intercept & $-0.423$ & --- & --- & $< 0.001$ \\
CoT (vs.\ Direct) & $+0.320$ & 1.38 & [1.22, 1.55] & $< 0.001$ \\
ICG-Self (vs.\ Direct) & $+0.319$ & 1.38 & [1.23, 1.53] & $< 0.001$ \\
Oracle (vs.\ Direct) & $+2.747$ & 15.59 & [11.66, 20.85] & $< 0.001$ \\
Claude Opus (vs.\ GPT-4o) & $+0.843$ & 2.32 & [2.04, 2.65] & $< 0.001$ \\
Gemini Flash (vs.\ GPT-4o) & $+0.284$ & 1.33 & [1.19, 1.48] & $< 0.001$ \\
Qwen3 (vs.\ GPT-4o) & $+0.167$ & 1.18 & [1.08, 1.30] & $= 0.001$ \\
\bottomrule
\end{tabular}
\end{table}

All condition effects remain highly significant after accounting for within-case correlation. The Oracle condition shows an odds ratio of 15.6$\times$ relative to Direct, consistent with the main Direct--Oracle comparison under the nested data structure.

\subsection{Cluster Bootstrap}

We perform case-level bootstrap resampling ($B = 10{,}000$): at each iteration, we resample 461 cases with replacement and compute condition-level recall within each bootstrap sample. This provides confidence intervals that respect the clustering structure.

\begin{table}[h]
\centering
\caption{Cluster bootstrap (resampling cases, $B = 10{,}000$): condition means and pairwise differences vs.\ Direct.}
\begin{tabular}{lrrr}
\toprule
\textbf{Comparison} & \textbf{$\Delta$ Recall} & \textbf{95\% CI} & \textbf{Significant} \\
\midrule
CoT vs.\ Direct & $+0.058$ & [0.030, 0.086] & Yes \\
ICG-Self vs.\ Direct & $+0.070$ & [0.044, 0.095] & Yes \\
Oracle vs.\ Direct & $+0.473$ & [0.441, 0.504] & Yes \\
\bottomrule
\end{tabular}
\end{table}

Both analyses give the same qualitative ordering as the main text, with confidence intervals that account for within-case clustering.

\subsection{Multiple Comparisons}

Our headline comparisons (Direct vs.\ Oracle, and the Reason/Region/Type decomposition) involve a small number of pre-specified contrasts with very large effect sizes ($p < 10^{-8}$ for the Oracle gap), which survive any standard correction. The per-subtype analysis is more exposed: it spans 19 subtypes, several with $N \leq 21$, so we treat the fine-grained subtype ordering as exploratory rather than confirmatory (see main-text Limitations) and rely on the category-level aggregation (5 cells, all $N \geq 56$) for inferential claims. We do not apply a formal family-wise correction to the exploratory subtype contrasts; readers should interpret individual mid-spectrum subtype differences accordingly.

\section{Cross-Judge Robustness Analyses}
\label{app:judge_breakdown}

We validate the primary Claude Sonnet 4.6 judge with two independent judges to test whether the headline findings depend on the choice of judge:
\begin{itemize}
    \item \textbf{GPT-4o} (in-pool cross-judge): shares the evaluated-model pool but uses identical judge prompts. We report agreement broken down by condition, category, evaluated model, and guided conditions (Sections M.1--M.4).
    \item \textbf{DeepSeek-V4-flash} (fully out-of-pool): neither evaluated, used for adversarial generation, nor used for pre-screening. We re-judge the full dataset under this independent instrument and report per-condition agreement (Sections M.5--M.6).
\end{itemize}
\noindent We then address two potential confounds: whether data source introduces selection bias (Section M.7) and whether the pattern generalizes to open-weight models (Section M.8).

Overall agreement between the primary Claude judge and GPT-4o is 84.2\% ($\kappa=0.675$) across the four primary conditions (12{,}416 paired constraint-level decisions). Under GPT-4o judging, Direct recall (macro-averaged) is 0.343 and Oracle recall is 0.921, yielding a Direct$\rightarrow$Oracle gap of $+0.578$. Per-condition breakdowns follow.

\subsection{Agreement by Condition}

\begin{table}[h]
\centering
\caption{Cross-judge agreement by prompting condition.}
\begin{tabular}{lrr}
\toprule
\textbf{Condition} & \textbf{Agreement} & \textbf{$\kappa$} \\
\midrule
Direct & 82.6\% & 0.638 \\
CoT & 80.9\% & 0.617 \\
ICG-Self & 82.4\% & 0.646 \\
Oracle & 91.0\% & 0.456 \\
\bottomrule
\end{tabular}
\end{table}

The Oracle condition shows the highest raw agreement (91\%) but the lowest $\kappa$ (0.456). This is a known property of Cohen's $\kappa$: when the base rate is extreme (Oracle recall $\approx$ 93\%, so nearly all decisions are ``found''), chance agreement $p_e$ is high, which deflates $\kappa$ even when raw agreement is excellent. The non-oracle conditions show consistent $\kappa \approx 0.62$--$0.65$, indicating moderate-to-substantial agreement on the more diagnostically relevant open-ended conditions.

\subsection{Agreement by Constraint Category}

\begin{table}[h]
\centering
\caption{Cross-judge agreement by constraint category.}
\begin{tabular}{lrr}
\toprule
\textbf{Category} & \textbf{Agreement} & \textbf{$\kappa$} \\
\midrule
Physical & 87.9\% & 0.751 \\
Cross-modal & 85.6\% & 0.712 \\
Semantic & 83.4\% & 0.610 \\
Style & 81.0\% & 0.611 \\
Spatial & 80.1\% & 0.574 \\
\bottomrule
\end{tabular}
\end{table}

Physical and Cross-modal constraints show the highest agreement ($\kappa > 0.7$), likely because the criteria for ``found'' are more clear-cut (e.g., did the model mention the reflection or not). Spatial constraints show the lowest $\kappa$ (0.574), reflecting ambiguity in whether a model ``sufficiently'' identified mirror/symmetry relationships versus merely mentioned the affected region. Style constraints ($\kappa = 0.611$) are similarly borderline---judges may disagree on whether a vague mention of ``maintain font consistency'' counts as identifying a specific font-matching constraint. Per-subtype conclusions involving Spatial and Style constraints should be interpreted with this higher noise floor in mind.

\subsection{Agreement by Evaluated Model}

\begin{table}[h]
\centering
\caption{Cross-judge agreement by evaluated model.}
\begin{tabular}{lrr}
\toprule
\textbf{Model} & \textbf{Agreement} & \textbf{$\kappa$} \\
\midrule
GPT-4o & 86.0\% & 0.719 \\
Claude Opus 4.6 & 85.0\% & 0.647 \\
Gemini 2.5 Flash & 83.3\% & 0.659 \\
Qwen3-VL-Plus & 82.6\% & 0.647 \\
\bottomrule
\end{tabular}
\end{table}

Agreement is highest for GPT-4o responses ($\kappa = 0.719$), possibly because GPT-4o's output style is more explicit in listing constraints, making judge decisions easier. The lower $\kappa$ for Claude ($0.647$) does not indicate self-preference bias---the Claude judge's positive rate on Claude responses exceeds its rate on GPT-4o responses by only $+0.016$---but may reflect Claude's tendency toward longer, more discursive responses where constraint identification is less clearly delineated.

\subsection{Extended Cross-Judge: Guided Conditions}

The cross-judge analysis above covers the four main conditions (Direct, CoT, ICG-Self, Oracle). We extended validation to the two guided self-discovery conditions (Taxonomy-Only, NL-Self) using a stratified 84-case sample (1,206 constraint-level judgments; one constraint-level decision was excluded due to a missing model response, yielding 603 paired judgments per condition).

\begin{table}[h]
\centering
\small
\caption{Extended cross-judge validation for guided conditions (84-case stratified sample, all 4 models). Claude Sonnet 4.6 as primary judge, GPT-4o as cross-judge.}
\setlength{\tabcolsep}{4pt}
\begin{tabular}{@{}lcccc@{}}
\toprule
\textbf{Condition} & \textbf{Claude} & \textbf{GPT-4o} & \textbf{Agree.} & \textbf{$N$} \\
\midrule
NL-Self & 0.635 & 0.620 & 78.6\% & 603 \\
Taxonomy-Only & 0.745 & 0.353 & 56.9\% & 603 \\
\midrule
Overall & 0.690 & 0.487 & 67.7\% & 1,206 \\
\bottomrule
\end{tabular}
\end{table}

NL-Self shows near-identical recall under both judges (0.635 vs.\ 0.620), consistent with it being a stable free-form self-discovery condition.
Taxonomy-Only is more judge-sensitive in this 84-case GPT-4o pilot: the primary Claude judge credits more borderline checklist-style matches than GPT-4o (0.745 vs.\ 0.353). This pilot serves as a sensitivity check. The full out-of-pool DeepSeek re-judging in Appendix~\ref{app:third_judge_taxonomy} gives the main Taxonomy-Only robustness result: DeepSeek assigns lower absolute recall than Claude (0.601 vs.\ 0.744), and the broad pattern is preserved---Claude and Qwen remain the top two models under both judges, although GPT-4o and Gemini swap positions under DeepSeek. Under both judges, Taxonomy-Only leaves a large gap to Oracle.

\paragraph{Signal-Guided cross-judge.} Signal-Guided (SG) is the strongest non-oracle condition outside Taxonomy-Only on average, so we run the same independent-judge test on the identical seed-42 stratified 100-case sample. Each SG constraint decision is re-judged by both the primary Claude Sonnet 4.6 judge and the non-Claude GPT-4o judge under one identical rubric (Table~\ref{tab:sg_crossjudge}).

\begin{table}[h]
\centering
\small
\caption{Signal-Guided cross-judge validation (seed-42 stratified 100-case sample). Claude Sonnet 4.6 primary vs.\ GPT-4o cross-judge under one identical rubric, broken out per generator model. $\Delta$ = Claude $-$ GPT-4o positive rate.}
\label{tab:sg_crossjudge}
\setlength{\tabcolsep}{4pt}
\begin{tabular}{@{}lccccc@{}}
\toprule
\textbf{Generator} & \textbf{Claude} & \textbf{GPT-4o} & \textbf{$\Delta$} & \textbf{Agree.} & \textbf{$N$} \\
\midrule
Claude Opus 4.6 & 0.593 & 0.651 & $-0.058$ & 86.0\% & 172 \\
GPT-4o          & 0.605 & 0.401 & $+0.204$ & 75.0\% & 172 \\
Gemini 2.5 Flash & 0.715 & 0.610 & $+0.105$ & 80.2\% & 172 \\
Qwen3-VL-Plus   & 0.593 & 0.465 & $+0.128$ & 75.6\% & 172 \\
\midrule
\textbf{All}    & \textbf{0.626} & \textbf{0.532} & $\mathbf{+0.094}$ & \textbf{79.2\%} & \textbf{688} \\
\bottomrule
\end{tabular}
\end{table}

Across the 688 SG decisions, the two judges agree on 79.2\% of decisions ($\kappa = 0.577$), with an overall positive-rate gap of $+0.094$. SG sits between judge-stable NL-Self and judge-sensitive Taxonomy-Only: guided prompting improves recall but does not approach Oracle and does not replace case-specific constraint information.

\subsection{Out-of-Pool Third Judge: Full Core-Condition Re-Judging}
\label{app:third_judge}

The cross-judge analysis above uses GPT-4o as a second judge. However, GPT-4o is not fully independent of the measured pool: it is itself one of the four \emph{evaluated} models, and it is also the model used to generate the adversarial constraints. To test whether the headline ordering could be an artifact of a judge model that overlaps the evaluated or generation pool, we re-judge the complete 461-case dataset with DeepSeek-V4-flash, a model that is \emph{neither evaluated, used for adversarial generation, nor used for candidate pre-screening}---a genuinely out-of-pool instrument. We use the \emph{identical} judge prompt across all four primary conditions; only the judge model differs (Table~\ref{tab:third_judge}).

\begin{table}[h]
\centering
\small
\caption{Out-of-pool third judge (DeepSeek-V4-flash) vs.\ primary Claude judge on the full 461-case dataset, all 4 models pooled. Recall and per-condition agreement ($\kappa$) are reported for each condition. The Direct$\rightarrow$Oracle gap reproduces under the independent judge; non-oracle self-discovery conditions remain substantially lower than Oracle.}
\label{tab:third_judge}
\setlength{\tabcolsep}{4pt}
\begin{tabular}{@{}lcccc@{}}
\toprule
\textbf{Condition} & \textbf{Claude} & \textbf{DeepSeek} & \textbf{$\kappa$} & \textbf{$N$} \\
\midrule
Direct   & 0.447 & 0.499 & 0.698 & 3{,}104 \\
CoT      & 0.527 & 0.597 & 0.622 & 3{,}104 \\
ICG-Self & 0.527 & 0.416 & 0.598 & 3{,}104 \\
Oracle   & 0.921 & 0.971 & 0.267 & 3{,}104 \\
\midrule
\textbf{All} & 0.606 & 0.621 & \textbf{0.675} & 12{,}416 \\
\bottomrule
\end{tabular}
\end{table}

Across 12{,}416 paired constraint-level decisions, the two judges agree with overall Cohen's $\kappa = 0.675$ (substantial), with near-identical positive rates (Claude 60.6\% vs.\ DeepSeek 62.1\%). The headline ordering remains stable: ordinary prompting and self-discovery remain well below Oracle, although DeepSeek scores ICG-Self lower than Direct.
The Direct$\rightarrow$Oracle gap also reproduces under the independent judge at the same node-level scoring granularity (DeepSeek: approximately $+0.47$; primary Claude judge on the same table: $+0.475$). Oracle has lower $\kappa$ (0.267) because both judges score it near ceiling, leaving little variance for chance-corrected agreement. Raw agreement remains high, and the headline gap does not depend on using a judge from the evaluated-model family.

\subsection{Taxonomy-Only Under Full Out-of-Pool Judging}
\label{app:third_judge_taxonomy}

We re-judge the full Taxonomy-Only set with DeepSeek-V4-flash using the identical judge prompt. The paired analysis contains 3{,}104 constraint-level decisions over the full 461-case set (Table~\ref{tab:third_judge_taxonomy}).

\begin{table}[h]
\centering
\small
\caption{Taxonomy-Only re-judging with out-of-pool DeepSeek-V4-flash (full 461-case set, all 4 models). Values are constraint-level recall (node-level positive rate), not case-level macro recall; the corresponding case-level macro values appear in Table~1 of the main text. DeepSeek assigns lower absolute recall but the broad model pattern is preserved, and Taxonomy-Only does not approach Oracle recall under either judge.}
\label{tab:third_judge_taxonomy}
\setlength{\tabcolsep}{5pt}
\begin{tabular}{@{}lcc@{}}
\toprule
\textbf{Model} & \textbf{Claude recall} & \textbf{DeepSeek recall} \\
\midrule
Claude Opus 4.6   & 0.898 & 0.827 \\
GPT-4o            & 0.701 & 0.508 \\
Qwen3-VL-Plus     & 0.726 & 0.554 \\
Gemini 2.5 Flash  & 0.649 & 0.515 \\
\midrule
\textbf{Average}  & \textbf{0.744} & \textbf{0.601} \\
\bottomrule
\end{tabular}
\end{table}

DeepSeek scores Taxonomy-Only lower than the primary judge (0.601 vs.\ 0.744). The broad pattern is preserved---Claude and Qwen remain the top two models under both judges---although GPT-4o and Gemini swap positions under DeepSeek. Under both judges, Taxonomy-Only is about 0.37 points below Oracle (DeepSeek: 0.601 vs.\ 0.971), showing that generic type reminders help models search more broadly but do not replace case-specific affected regions or causal dependency reasons.

\paragraph{Judge sensitivity is condition-specific.}
Table~\ref{tab:percond_judge} reports per-condition agreement between the primary Claude judge and DeepSeek on the full dataset. Open-ended non-oracle conditions show substantial agreement ($\kappa = 0.60$--$0.70$) with small positive-rate differences. Oracle $\kappa$ is deflated by base-rate effects because both judges score it near ceiling. Taxonomy-Only has a larger positive-rate gap ($+0.143$), yet the broad model pattern is preserved and the condition does not close the discovery gap under either judge.

\begin{table}[h]
\centering
\small
\caption{Per-condition agreement between the primary Claude judge and out-of-pool DeepSeek judge, on the full 461-case set across all conditions. ``Pos.\ gap'' = Claude positive rate minus DeepSeek's.}
\label{tab:percond_judge}
\setlength{\tabcolsep}{4pt}
\begin{tabular}{@{}lccc@{}}
\toprule
\textbf{Condition} & \textbf{$\kappa$} & \textbf{Pos.\ gap} & \textbf{$N$} \\
\midrule
Direct        & 0.698 & $-0.052$ & 3{,}104 \\
CoT           & 0.622 & $-0.070$ & 3{,}104 \\
ICG-Self      & 0.598 & $+0.111$ & 3{,}104 \\
Oracle        & 0.267$^\dagger$ & $-0.050$ & 3{,}104 \\
\midrule
Taxonomy-Only & 0.594 & $+0.143$ & 3{,}104 \\
\bottomrule
\end{tabular}
\end{table}

\noindent$^\dagger$Oracle $\kappa$ is deflated by base-rate effects: both judges score Oracle near ceiling ($\geq$0.92), so variance is low and $\kappa$ underestimates true agreement; raw agreement is $>$90\%.

\subsection{Source-Stratified Analysis}
\label{app:source_stratified}

To assess whether candidate pre-screening by Claude Opus 4.6 introduced selection bias, we stratified Direct recall and constraint-subtype distribution by data source.

\begin{table}[h]
\centering
\small
\caption{Direct recall by data source and model (Claude Sonnet judge). Parentheses after source names indicate case counts; the $N$ column reports constraint-node counts. Model ranking is preserved across all sources.}
\label{tab:source_recall}
\setlength{\tabcolsep}{3.5pt}
\begin{tabular}{@{}lccccr@{}}
\toprule
\textbf{Source (cases)} & \textbf{Claude} & \textbf{GPT-4o} & \textbf{Gemini} & \textbf{Qwen3} & \textbf{Nodes} \\
\midrule
Other/Manual (139) & 0.681 & 0.383 & 0.441 & 0.394 & 263 \\
Pexels (215) & 0.679 & 0.295 & 0.460 & 0.411 & 350 \\
TextEditBench (107) & 0.590 & 0.287 & 0.463 & 0.437 & 163 \\
\bottomrule
\end{tabular}
\end{table}

\noindent Source labels in this diagnostic table follow the metadata categories used by the stratification script rather than the original hosting platform for every collected image. Singleton source labels with insufficient sample size are folded into ``Other/Manual,'' which includes Unsplash, Canva, and images collected through the \texttt{find\_data/} pipeline. With this grouping, the table covers the full dataset (461 cases, 776 nodes).

Claude's Direct recall on TextEditBench-sourced cases (0.590) is \emph{lower} than on other sources (0.679--0.681), providing no support for a pre-screening inflation hypothesis. The model ranking (Claude $>$ Gemini $>$ Qwen3 $>$ GPT-4o) is preserved across all three source groups.

We additionally verify that subtype distribution is not dominated by any single source (Table~\ref{tab:source_subtype}).

\begin{table}[h]
\centering
\footnotesize
\caption{Constraint-category distribution by data source (\% of nodes from that source). All five categories are represented in every source; TextEditBench is more style/semantic-heavy while Pexels and Other/Manual skew toward physical and cross-modal.}
\label{tab:source_subtype}
\setlength{\tabcolsep}{2pt}
\begin{tabular}{@{}lrrrrr@{}}
\toprule
\textbf{Source} & \textbf{Phys.} & \textbf{Spat.} & \textbf{Sem.} & \textbf{Style} & \textbf{Cross-mod.} \\
\midrule
Other/Manual (263 nodes)  & 28\% & 7\%  & 25\% & 12\% & 28\% \\
Pexels (350 nodes)        & 23\% & 8\%  & 15\% & 25\% & 29\% \\
TextEditBench (163 nodes) & 10\% & 5\%  & 28\% & 33\% & 24\% \\
\bottomrule
\end{tabular}
\end{table}

\noindent All five categories are represented in every major source group shown; no top-level category is exclusive to a single source. The style-heavy skew of TextEditBench reflects that dataset's focus on typographic and design images, but all five categories remain represented, and the recall ordering from Table~\ref{tab:source_recall} holds within each source group.

\subsection{Open-Weight Generalization}
\label{app:open_weight}

Our four primary models are all proprietary, which leaves open whether local-edit bias reflects a property of closed models or a single training lineage. To test this, we evaluate GLM-4.6V---an open-weight vision-language model from an independent developer (distinct from the four evaluated families)---under the Direct and Oracle conditions, using the \emph{identical} prompts and the same Claude Sonnet judge. We run it on a 50-case stratified subset, reporting the proprietary models on the same subset for comparison (Table~\ref{tab:open_weight}).

\begin{table}[h]
\centering
\small
\caption{Direct vs.\ Oracle recall on the 50-case stratified subset (Claude Sonnet judge, macro-averaged). The open-weight GLM-4.6V falls within the range spanned by the proprietary models on this subset, showing the same qualitative Direct$\rightarrow$Oracle pattern.}
\label{tab:open_weight}
\setlength{\tabcolsep}{6pt}
\begin{tabular}{@{}llcc r@{}}
\toprule
\textbf{Model} & \textbf{Type} & \textbf{Direct} & \textbf{Oracle} & \textbf{Gap} \\
\midrule
Claude Opus 4.6 & proprietary & 0.650 & 0.953 & $+0.303$ \\
GPT-4o & proprietary & 0.309 & 0.996 & $+0.687$ \\
Gemini 2.5 Flash & proprietary & 0.408 & 0.928 & $+0.520$ \\
Qwen3-VL-Plus & proprietary & 0.391 & 0.926 & $+0.535$ \\
\midrule
GLM-4.6V & open-weight & 0.388 & 0.970 & $+0.582$ \\
\bottomrule
\end{tabular}
\end{table}

\noindent GLM-4.6V's Direct recall (0.388), Oracle recall (0.970), and gap ($+0.582$) all lie within the proprietary range on this subset (Direct 0.31--0.65, Oracle 0.93--1.00). The same discovery-stage pattern appears: GLM-4.6V misses many constraints under Direct and recovers most of them when they are made explicit. This extends the local-edit bias observation beyond the four proprietary models; the 50-case, single-model setting makes the result suggestive rather than conclusive.

\section{Additional Discussion}
\label{app:additional_discussion}

\subsection{What Drives Discoverability: Signal Salience and Relation Inferability}

We propose that constraint discoverability depends on two interacting factors:
\begin{enumerate}
    \item \textbf{Edit signal salience}: Does the edit action itself strongly cue that other regions may need updating? Changing a number (``50\% $\rightarrow$ 30\%'') immediately signals that derived quantities may be affected. Changing decorative text (``GOOD MORNING $\rightarrow$ GOOD EVENING'') does not obviously signal a scene-level temporal mismatch.
    \item \textbf{Relation inferability}: Given that a dependency exists, can the model infer it from local information, or does it require cross-region visual search or external knowledge? Mathematical relations are locally inferable from the edit itself; cultural coherence requires world knowledge that the edit does not carry.
\end{enumerate}

\noindent To operationalize this framework for correlation analysis, we assign each subtype a single \emph{inference indirectness} score that composites both factors into an ordinal scale: \textbf{Level~0} = dependency directly computable from the edit itself (high signal, high inferability; e.g., mathematical derivation); \textbf{Level~1} = dependency requires noticing a direct spatial or visual relationship with the edit target (moderate signal; e.g., its reflection or shadow); \textbf{Level~2} = dependency requires cross-region search or indirect reasoning (low signal or low inferability; e.g., a spatially distant symmetric counterpart); \textbf{Level~3} = dependency requires external world knowledge not present in the image (low signal and low inferability; e.g., cultural conventions).

On the full 461-case set, this coding correlates with per-subtype Direct recall at Spearman $\rho = -0.53$ ($p = 0.020$)---the value cited in the main text---with a monotonic pattern across levels: Level~0 averages 0.800, Level~1 0.492, Level~2 0.414, and Level~3 0.181. (On the smaller original non-exploratory subtype subset the same coding gives a stronger $\rho = -0.753$, $p = 0.007$.)

However, this single-axis coding is coarse: two subtypes at the same level can differ substantially in recall when their signal--inferability profiles diverge. For example, Level~1 includes both reflection (0.69, high signal from visible reflective surface) and style-color (0.41, low signal because text changes do not cue palette rules). The two-factor decomposition explains this within-level variation:
\begin{itemize}
    \item \textbf{Numerical (0.87)}: Signal strong (numeric change is salient) + relation inferable (mathematical).
    \item \textbf{Visual-data (0.73)}: Signal strong (label/title change) + relation inferable (chart-data mapping is direct).
    \item \textbf{Temporal---date/weekday (high)}: Signal strong (date change) + relation inferable (calendar logic).
    \item \textbf{Temporal---scene-level (low)}: Signal weak (text change does not cue scene mismatch) + relation requires contextual reasoning.
    \item \textbf{Cultural (0.18)}: Signal weak (text content change carries no cultural flag) + relation requires external knowledge.
    \item \textbf{Occlusion (0.34)}: Signal weak (edit target does not cue hidden regions) + relation requires spatial search beyond the edit region.
    \item \textbf{Style-color (0.41)}: Signal weak (text change does not cue palette rules) + relation ambiguous (is the color scheme binding or decorative?).
\end{itemize}

\noindent We treat this as an explanatory hypothesis supported by the subtype pattern, not as a validated causal model.
For system design, interventions should target the corresponding failure source: low-signal constraints require explicit attention guidance, while low-inferability constraints require external knowledge retrieval or domain-specific reasoning scaffolds.

\subsection{Error Taxonomy}

Looking at Oracle failures, we identified three preliminary patterns based on a manually inspected sample of 21 failure instances:
\begin{enumerate}
    \item \emph{Visual grounding failures} (9/21, 43\%)---the model acknowledges the constraint but cannot locate the affected region.
    \item \emph{Constraint interpretation failures} (7/21, 33\%)---the model misreads what the constraint requires (e.g., confusing color-matching with font-matching).
    \item \emph{Instruction override} (5/21, 24\%)---the model's instruction-following behavior wins out when it perceives a conflict with the constraint.
\end{enumerate}
Each category points to a different fix: better visual grounding for (1), structured reasoning for (2), and constraint-aware instruction tuning for (3).

\subsection{Practical Directions}

Two applications follow from our results.
First, automated constraint elicitation: a module that, given $(I, T)$, proposes a candidate \ours{} for human review before the edit runs.
The 87.3\% precision of ICG-Self-generated constraints makes this feasible with moderate oversight.
Second, constraint-guided generation: feeding the oracle graph as structured conditioning during image synthesis, much as ControlNet~\citep{zhang2023adding} conditions on spatial structure.
The near-ceiling Oracle recall (0.938) shows that models can \emph{use} constraints well once they have them.

\subsection{Edit Planning Experiment}
\label{app:planning}

\paragraph{Setup.}
We sample 60 cases stratified by constraint category (12 per category) from the full 461-case set.
For each case we identify up to 2 ground-truth constraints that require a visible, independently verifiable plan step, yielding 100 ground-truth constraints total.
Each case is evaluated under three conditions: \textbf{Direct} (no constraint guidance), \textbf{NL-Self} (model's own discovered constraints), and \textbf{Oracle} (ground-truth constraints provided verbatim).
In all three conditions, Claude Opus 4.6 is prompted to produce a step-by-step text edit plan given the image and instruction.

\paragraph{Judge.}
Claude Sonnet 4.6 scores each ground-truth constraint as satisfied (1) or not (0) given the produced plan.
Plan satisfaction = fraction of ground-truth constraints satisfied.
The judge uses the same rubric as the core recall judge, adapted to assess plan coverage rather than discovery.

\paragraph{Results.}
Table~\ref{tab:planning_results} reports plan satisfaction under each condition.

\begin{table}[h]
\centering
\small
\caption{Edit planning experiment (60 stratified cases, 100 ground-truth constraints). Plan satisfaction is the fraction of ground-truth constraints covered by the generated edit plan.}
\label{tab:planning_results}
\begin{tabular}{@{}lcc@{}}
\toprule
\textbf{Condition} & \textbf{Plan satisfaction} & \textbf{Count} \\
\midrule
Direct & 0.700 & 70/100 \\
NL-Self & 0.680 & 68/100 \\
Oracle & 0.990 & 99/100 \\
\bottomrule
\end{tabular}
\end{table}

\noindent Oracle significantly improves over Direct ($\Delta = 0.29$, $p < 10^{-8}$, McNemar), whereas NL-Self does not improve over Direct ($p = 0.62$).

\subsection{Two-Stage Discover-then-Verify Pipeline}
\label{app:two_stage}

A two-stage verifier tests whether post-hoc filtering can turn self-discovered constraints into oracle-quality planning input. We run this pipeline on a related 58-case set.

\paragraph{Setup.}
\emph{(i) Discover:} take the NL-Self response (Claude Opus 4.6) and extract up to 10 candidate constraints per case.
\emph{(ii) Verify:} Claude Opus 4.6 visually inspects each candidate and keeps only those whose region is identifiable, whose dependency is observable in the image, and whose update is necessary.
\emph{(iii) Plan:} Claude Opus 4.6 produces an edit plan from the verified set.
\emph{(iv) Judge:} Claude Sonnet 4.6 scores each ground-truth constraint as satisfied or not, identical to the main-text planning judge.
Across the 58 cases the verifier removed 181 of 580 proposals (31.2\%), showing it is active rather than a pass-through.

\paragraph{Result.}
Plan satisfaction is 0.712 (74/104 ground-truth constraints), numerically close to Direct planning (0.700) and much lower than Oracle's 0.990.
Filtering the model's own candidates does not approach the quality obtainable from ground-truth constraints; the limiting factor is generating the right dependencies at discovery time, not pruning false positives afterward.

\paragraph{Comparability.}
The 58-case pipeline is a related diagnostic run, not a controlled comparison to the 60-case single-pass planning experiment. It differs in input representation, planner prompt, and generation budget. We therefore treat it as suggestive evidence rather than a formal paired test: even after active filtering, self-discovered constraints do not approach plans based on ground-truth constraints.

\subsection{Precision Details}

Precision analysis on ICG-Self (JSON format) shows 87.3\% precision (6628 valid / 7594 total constraint predictions across 461 cases $\times$ 4 models).
NL-Self (free-form) achieves 83.1\% precision (10553 valid / 12701 total).
Per-model precision for ICG-Self: Gemini 88.9\%, Claude 88.3\%, Qwen3 86.8\%, GPT-4o 84.2\%.
For NL-Self: Gemini 87.4\%, Qwen3 85.0\%, Claude 84.3\%, GPT-4o 73.1\%.

The recall-precision tradeoff reveals the core issue: models achieve high precision but low recall.
NL-Self produces more predictions per case than ICG-Self with similar precision, explaining its higher recall.
The dominant failure mode is omission rather than hallucination: models fail to notice constraints that are visually present.


\section{Human Baseline Protocol}
\label{app:human_baseline}

Three annotators with computer-science backgrounds---not involved in dataset construction or taxonomy design---independently completed the same task as the models under the Direct condition: given an image and editing instruction, list all considerations for maintaining consistency after the edit. They received no taxonomy checklist, no ground-truth access, and no inter-annotator discussion.

Their free-form responses were scored by the same Claude Sonnet 4.6 judge under the identical rubric applied to model outputs. The expanded human baseline consists of 100 cases selected via stratified sampling.

\textbf{Results} (micro-averaged recall, $N = 173$ constraints, 100-case stratified sample):

\begin{center}
\begin{tabular}{@{}lc@{}}
\toprule
\textbf{Annotator} & \textbf{Recall} \\
\midrule
A1 & 0.64 \quad (111/173) \\
A2 & 0.68 \quad (118/173) \\
A3 & 0.70 \quad (121/173) \\
\textbf{Mean} & \textbf{0.67} \\
\bottomrule
\end{tabular}
\end{center}

The mean recall (0.67) exceeds unguided models under comparable node-level scoring (0.447 micro; 0.461 case-macro in Table~1 of the main paper), indicating higher spontaneous recall for attentive humans than for current models under this rubric. Human recall remains below Oracle-level performance under comparable node-level scoring (0.921 micro; 0.938 case-macro): even the strongest annotator missed roughly 30\% of constraints, so spontaneous discovery is difficult for both humans and models.

\section{OCR-Assisted Prompting}
\label{app:ocr_assisted}

We test whether providing an explicit OCR transcript of all visible text in the image reduces the discovery gap by reducing visual text-reading difficulty as a possible confound.

\subsection{Setup}

We sample a stratified 50-case subset (10 per constraint category) using seed~42 from the full 461-case set. For each case, we first extract all visible text via a dedicated OCR pass (GPT-4o, prompted to list every text element with its approximate location and visual context). The resulting transcript is prepended to the standard discovery prompt, and the model is shown the image alongside the OCR text. All four models (Claude Opus~4.6, GPT-4o, Gemini 2.5~Flash, Qwen3-VL-Plus) are evaluated under the OCR-assisted condition; responses are judged by the same Claude Sonnet~4.6 judge used for the core conditions.

\subsection{Results}

\begin{table}[h]
\centering
\small
\caption{OCR-assisted prompting: per-model macro-averaged recall on the 50-case stratified subset, compared with Direct and Oracle recall on the same subset. The All row is pooled over all case--model decisions rather than the unweighted mean of the four per-model values.}
\label{tab:ocr_assisted}
\setlength{\tabcolsep}{5pt}
\begin{tabular}{@{}lcccc@{}}
\toprule
\textbf{Model} & \textbf{Direct} & \textbf{OCR-Assisted} & \textbf{Oracle} & \textbf{$\Delta$ (OCR$-$Dir.)} \\
\midrule
Claude Opus 4.6   & 0.650 & 0.763 & 0.953 & $+0.113$ \\
GPT-4o            & 0.309 & 0.702 & 0.996 & $+0.393$ \\
Gemini 2.5 Flash  & 0.408 & 0.673 & 0.928 & $+0.265$ \\
Qwen3-VL-Plus     & 0.391 & 0.701 & 0.926 & $+0.310$ \\
\midrule
\textbf{All}      & 0.436 & \textbf{0.710} & 0.945 & $+0.274$ \\
\bottomrule
\end{tabular}
\end{table}

OCR-assisted prompting improves recall over Direct ($\Delta = +0.274$ macro-averaged), showing that text reading contributes to the gap. GPT-4o benefits most ($+0.393$), consistent with its low Direct recall on visually complex text. A substantial gap to Oracle remains ($\Delta = -0.235$): models must still infer which visible text elements are causally linked to the edit and why. OCR-assisted retrieval reduces text-reading difficulty, but it does not replace that inference step.

\section{Evaluation Details}
\label{app:eval_details}

\subsection{Parameters and Preprocessing}

All model calls used a temperature of $0.0$ and \texttt{max\_tokens} set to $4096$ to minimize decoding randomness and prevent truncation, with the exception of Gemini 2.5 Flash which used \texttt{max\_tokens = 16384} during chain-of-thought conditions to accommodate its long internal reasoning steps.

Before encoding, all input images were standardized to a maximum long-edge resolution of 1024 pixels using bilinear interpolation while strictly preserving the original aspect ratio. Standard base64-encoded strings were then embedded directly into the multimodal user message payload.

For API robustness, requests were managed through a wrapper with a maximum of 3 retries and an exponential backoff starting at 5 seconds. Failed runs were logged and rerun until completion.

\bibliography{aaai2026}